# A thermodynamically consistent physics-informed deep learning material model for short fiber/polymer nanocomposites

Betim Bahtiri[a,*], Behrouz Arash[b], Sven Scheffler[a], Maximilian Jux[c], Raimund Rolfes[a]

[a]*Institute of Structural Analysis, Leibniz Universität Hannover, Appelstraße 9A, 30167 Hannover, Germany*
[b]*Department of Mechanical, Electrical and Chemical Engineering, Oslo Metropolitan University, Pilestredet 35, 0166 Oslo, Norway*
[c]*Institute of Lightweight Systems, Multifunctional Materials, DLR (German Aerospace Center), Lilienthalplatz 7, 38108 Brunswick, Germany*

**Abstract**

This work proposes a physics-informed deep learning (PIDL)-based constitutive model for investigating the viscoelastic-viscoplastic behavior of short fiber-reinforced nanoparticle-filled epoxies under various ambient conditions. The deep-learning model is trained to enforce thermodynamic principles, leading to a thermodynamically consistent constitutive model. To accomplish this, a long short-term memory network is combined with a feed-forward neural network to predict internal variables required for characterizing the internal dissipation of the nanocomposite materials. In addition, another feed-forward neural network is used to indicate the free-energy function, which enables defining the thermodynamic state of the entire system. The PIDL model is initially developed for the three-dimensional case by generating synthetic data from a classical constitutive model. The model is then trained by extracting the data directly from cyclic loading-unloading experimental tests. Numerical examples show that the PIDL model can accurately predict the mechanical behavior of epoxy-based nanocomposites for different volume fractions of fibers and nanoparticles under various hygrothermal conditions.

*Keywords:* Short fiber/Epoxy Nanocomposites, Physics-informed Neural Networks, Recurrent Neural Network, Thermodynamic Consistent Modeling, Finite Deformation

## 1. Introduction

In the coming decades, materials science is expected to make breakthroughs in order to meet the increasing demand for lightweight and durable materials that can withstand extreme conditions. The current focus of research in synthetic composites is on developing high-performance glass fiber (GF) reinforced epoxy composites that incorporate various nanofillers such as nanoparticles and nanotubes [1, 2]. Adding nanoscale fillers to polymer matrices has a dual impact on material behavior. Firstly, nanoparticles exhibit different physical and chemical properties compared to their bulk counterparts due to a more significant portion of their atoms being on the surface. Secondly, nanoparticles have a remarkably high surface-to-volume ratio that provides a contact surface area that can be up to 1000 times greater than that of micro-sized particles. This enhances efficient load transfer from the matrix to the reinforcements, which is critical for achieving high-performance materials. In recent studies, boehmite nanoparticles (BNPs) have been shown to be effective in enhancing the material properties of composites, such as the strength, critical energy release rate and other material properties [3–5].

*Corresponding author
*Email address:* b.bahtiri@isd.uni-hannover.de (Betim Bahtiri)

Although the resulting fiber-reinforced BNP/epoxy nanocomposites have proven to be suitable for different engineering applications, predicting their mechanical behavior under various loading conditions has become a real challenge. To accurately predict the history-dependent behavior of these materials under hygrothermal conditions, the literature has seen the development of more intricate constitutive models with additional parameters [6–10]. While these advanced constitutive models effectively capture the highly nonlinear nature of material behavior, a significant challenge lies in minimizing the error between numerical predictions and experimental data, particularly under complex loading conditions and different ambient conditions, such as temperature and moisture. Additionally, classical constitutive laws often lack generality and accuracy to capture the mechanical behavior under different ambient conditions [11–13].

In recent years, deep learning approaches have attracted attention to improving the numerical methods [14–18]. One of the first works in this direction was conducted by Ghaboussi et al. [19], who proposed a neural network approach to unify the mechanical behavior of plain concrete directly from experimental data. Recognizing that the material behavior is path-dependent, they trained the neural network to predict strain increments given the current state of stress, strain, and stress increment. However, due to the necessity for a comprehensive set of experiments, the model was limited to predicting only biaxial and uniaxial loading scenarios. Also, long short-term memory networks (LSTM), which belong to the class of recurrent neural networks (RNN), have been incorporated within finite element simulations [20] to computational homogenize anisotropic behavior under arbitrary loading paths [21] for het-erogeneous material. The results prove the effectiveness of the LSTM network in capturing the mechanical behavior for a two-dimensional case. The classical way of utilizing machine learning in constitutive modeling is to map the strain tensor and material parameters to depict the stress tensor by creating a set of data from constitutive models [22, 23]. Bahtiri et al. [24] presented a similar approach to predict the viscoelastic-viscoplastic damage of a nanoparticle-filled epoxy system under cyclic loading-unloading conditions with moisture. The authors developed a deep learning material model, implemented the model within a finite element framework and showed a significant increase in computational efficiency compared with the classical constitutive model.

While RNNs efficiently predict path-dependent mechanical behavior, the plain mapping of strain to stress tensors does not enforce thermodynamic consistency and can not guarantee accurate extrapolation. Therefore, using a physics-informed approach [25] has proven to be successful in various engineering applications, such as solving differential equations [25–27] and constitutive modeling [28–30]. Masi et al. [31] proposed a physics-informed neural network to predict strain rate-independent material behavior by encoding the two principles of thermodynamics within the neural network's architecture. They demonstrate the applicability of their model for elastoplastic materials. This work is extended to an evolution of their neural networks towards identifying internal variables for path-dependent materials and evolution equations by decoupling from their previous incremental formulation [32]. Wang et al. [33] presented a physics-informed constitutive neural network for quasi-linear viscoelasticity by combining an initial time-independent stress and a time-dependent relaxation using a recurrent network. The results prove that combining these two networks leads to an accurate prediction of soft viscoelastic tissues. Linden et al. [34] proposed a physics-informed neural network to enforce thermodynamic consistency and additional constitutive conditions for the hyperelastic behavior of a Neo-Hooke model. Their model is trained solely by using uniaxial stress data and shows accurate results for predicting biaxial or simple shear tests for anisotropic material. As'ad et al. [35] presented a different physics-informed neural network to predict complex viscoelastic behavior for a woven fabric material. They employ an internal-variable three-potential approach to viscoelasticity and satisfy the thermodynamics, stability conditions, consistency, and recovery of the elastic limits. The performance of their physics-informed deep learning (PIDL) model is illustrated on a computational homogenization simulation. Another work that utilizes the capabilities of recurrent neural networks is from He et al. [36], who presented a PIDL model to predict the internal variables of a path-dependent material. They predict the internal variables and the free-energy function by combining several neural networks. Their results are accurate even for cyclic shear loading using experimental stress-strain data of soil material. The latest work of Abdolazizi et al. [37] offers a constitutive artificial neural network model that captures the anisotropy of viscoelasticity at finite deformation. Their model relies on the concept of generalized Maxwell models, and the architecture of their PIDL model is restricted to predicting viscoelastic behavior. The authors train their model on stress-strain data from synthetic data of very-high-bond soft electro-active polymers and experimental stress-strain data of very-high-bond polymers, showing accurate interpolation and extrapolation behavior. Rosenkranz et al. [38] present a similar approach, which is based on internal variables and can predict nonlinear viscoelasticity by using a free-energy and dissipation potential in combination with input convex neural networks [39]. They utilize invariants and show the advantages of invariants as input instead of the coordinates themselves.

After reviewing the investigations mentioned in the literature, it is evident that several issues related to these models require attention. Firstly, the models are either based on small strain assumption or are proposed for a specific mechanical behavior, such as viscoelasticity [37]. Secondly, the models have mainly been developed for homogenous or anisotropic materials at specific temperatures and dry condition. This neglects the influence of different ambient conditions (i.e., temperature, and moisture content) in combination with other inputs such as particle or fiber volume fraction. Moreover, their predictive capability has been compared solely against synthetic data from classical constitutive models to increase computational efficiency [24] or demonstrate their applicability [32].

To address these issues, this study focuses on developing a PIDL model to improve the performance of constitutive modeling for short fiber/epoxy nanocomposite material under different ambient conditions. For this, we propose an architecture of a PIDL model for an anisotropic material system with multiple families of fibers at finite deformation. The proposed model can predict the internal variables and enforce the thermodynamic principles, thus increasing the generalization performance. The effectiveness of the proposed PIDL model is evaluated in two cases. The first case includes synthetic data from a classical viscoelastic-viscoplastic model for a fiber-reinforced nanoparticle-filled epoxy nanocomposite calibrated using experimental data with moisture content. The second case comprises experimental data for a nanoparticle-filled epoxy nanocomposite for a wide range of temperatures, including dry and saturated states. The accuracy and effectiveness of the present model are shown by evaluating its interpolation and extrapolation capabilities.

The present work is organized as follows. First, the thermodynamic principles of constitutive modeling are presented in Section 2, which are incorporated within the PIDL model. The integration and explanation of the model are provided in Section 3. We deliver insights into the neural networks used, the symmetry classes of the material at hand, and the overall architecture of the PIDL model. The training approach is also explained. In the last Section 4 we present our classical viscoelastic-viscoplastic damage model for fiber reinforced nanoparticle-filled epoxy nanocomposite. The model is calibrated using experimental data, and data is generated using the calibrated constitutive model to show the accuracy of the PIDL model for a three-dimensional case of an anisotropic material system with moisture content and different volume fractions of fibers/nanoparticles. Lastly, the predictive capability of the PIDL model is evaluated by simulating cyclic loading-unloading of a nanoparticle-filled epoxy nanocomposite using experimental data for various ambient conditions.

## 2. Thermodynamics principles and general requirements for constitutive modelling

The first law of thermodynamics states that the rate of change of total kinetic energy of a thermodynamic system equals the rate at which external mechanical work is done on the system at hand. Therefore, the local form and material description of the energy balance reads

$$\dot{e} = \mathbf{P} : \dot{\mathbf{F}} - \mathrm{Div}\mathbf{Q} + R, \tag{1}$$

where $e$ and $\dot{e}$ are the volume density of the internal energy and its local time derivative, $\mathbf{P}$ is the first Piola-Kirchhoff stress tensor, $\mathbf{F}$ is the deformation gradient, $\mathbf{Q}$ is the referential heat flux and $R$ is the volume density of possible external source terms. The second law of thermodynamics is responsible for the direction of the energy transfer process. The combination of the first and second law of thermodynamics and elimination of the external heat source R leads to the Clausius-Duhem inequality as follows [40]:

$$\mathbf{P} : \dot{\mathbf{F}} - \dot{e} + \theta\dot{\eta} - \frac{1}{\theta}\mathbf{Q} \cdot \mathbf{Grad}\theta \geq 0, \tag{2}$$

where the last term describes the entropy production by conduction of heat. By applying the Legendre transformation, the Helmholtz free energy density is expressed as $\psi = e - \theta\eta$ [41].

Whenever the temperature gradient is equal to 0, every thermodynamic system must fulfill the Clausius-Duhem inequality

$$\mathrm{d} = \mathbf{P} : \dot{\mathbf{F}} - \dot{\psi} - \dot{\theta}\eta \geq 0. \tag{3}$$

For the case of a purely mechanical case, the thermal effects can be ignored and the inequality reads

$$\mathrm{d} = \mathbf{P} : \dot{\mathbf{F}} - \dot{\psi} \geq 0, \tag{4}$$

and the free energy $\psi$ coincides with the internal energy $e$ as shown within the Legendre transformation.

In order to model nonlinear and inelastic materials, it is necessary to derive a constitutive model that is capable of describing every dissipative process. The thermodynamic state of materials, including dissipation, can be determined by a number of so-called internal variables, which can be scalar, vector, or tensor values. We denote the internal variables collectively by $z$. These variables describe the internal structure of the material at hand at each time, and the evolution of these variables replicates the history of the deformation.

*2.1. Constitutive equations and internal dissipation*

To describe the dissipation, the Helmholtz free-energy function $\psi$, which defines the thermodynamic state, is postulated as follows:

$$\psi = \psi(\mathbf{F}, z_1, \ldots, z_m). \tag{5}$$

To particularize the Clausius-Duhem inequality as described in Eq. (4) to the postulated free-energy $\psi$, we differentiate Eq. (5) with respect to time, leading to the following dissipation inequality:

$$\mathrm{d} = \left(\mathbf{P} - \frac{\partial \psi(\mathbf{F}, z_1, \ldots, z_m)}{\partial \mathbf{F}}\right) : \dot{\mathbf{F}} - \sum_{\alpha=1}^{m} \frac{\partial \psi(\mathbf{F}, z_1, \ldots, z_m)}{\partial z_\alpha} : \dot{z}_\alpha \geq 0. \tag{6}$$

To satisfy the dissipation inequality, the following must hold at every point of the continuum body at all times:

$$\mathbf{P} = \frac{\partial \bar{\psi}(\mathbf{F}, z_1, \ldots, z_m)}{\partial \mathbf{F}} \quad , \qquad \mathrm{D} = -\sum_{\alpha=1}^{m} \frac{\partial \psi(\mathbf{F}, z_1, \ldots, z_m)}{\partial z_\alpha} : \dot{z}_\alpha \geq 0 \quad , \tag{7}$$

The above equation restricts the free energy $\psi$ and relates the gradient of $\psi$ concerning the internal variables $z_\alpha$. In the Eq. (7), the deformation gradient and the internal variables are associated with the thermodynamic forces $\mathbf{P}$ and the internal variables $z$, respectively. These restrictions satisfy the fundamental Clausius-Duhem inequality characterizing the local entropy production as presented in Eq. (4). Besides the thermodynamic principles, additional requirements must be satisfied, such as objectivity, material symmetry, polyconvexity, non-negativity of the free energy, and normalization conditions. These are included within the PIDL model, as presented below.

### *2.2. Material symmetry classes for composites*

In the following, two symmetry classes are presented for composite materials. While particle-reinforced nanocomposites are isotropic materials, we present another symmetry class for fiber-reinforced and nanoparticle-filled epoxy nanocomposites. To ensure objectivity and material symmetry, the free-energy function is formulated in terms of invariants from the right Cauchy-Green deformation tensor $\mathbf{C} = \mathbf{F}^T \cdot \mathbf{F}$ leading to the following:

$$\psi = \psi(I_1(\mathbf{C}), ..., I_m(\mathbf{C}), z_1, \ldots, z_m). \tag{8}$$

The corresponding second Piola-Kirchhoff stress tensor is given by:

$$\mathbf{S} = 2\frac{\partial \psi}{\partial \mathbf{C}} = 2\sum_{\alpha=1}^{m} \frac{\partial \psi}{\partial I_\alpha}\frac{\partial I_\alpha}{\partial \mathbf{C}}, \tag{9}$$

which implicitly enforces the first thermodynamic principle and the first part of the Clausius-Duhem inequality from Eq. (7), while the second principle of the thermodynamics, i.e., $\mathrm{D} \geq 0$ is enforced within the loss function, which is discussed in the following subsection. Also, using the invariant-based approach, the symmetry condition of the second Piola-Kirchhoff stress tensor, objectivity, and symmetry of the material class at hand is fulfilled.

*Isotropic model*

For the isotropic material model, three invariants of the tensor $\mathbf{C}$ are, in general, sufficient to describe the stress state [40]. The invariants are defined as follows:

$$\begin{aligned} &I_1(\mathbf{C}) = \mathrm{trC} = \lambda_1^2 + \lambda_2^2 + \lambda_3^2 \ , \\ &I_2(\mathbf{C}) = \tfrac{1}{2}\left[(\mathrm{tr}\mathbf{C})^2 - \mathrm{tr}(\mathbf{C}^2)\right] = \mathrm{tr}\mathbf{C}^{-1}\mathrm{det}\mathbf{C} = \lambda_1^2\lambda_2^2 + \lambda_1^2\lambda_3^2 + \lambda_2^2\lambda_3^2 \ , \\ &I_3(\mathbf{C}) = \mathrm{det}\mathbf{C} = J^2 = \lambda_1^2\lambda_2^2\lambda_3^2, \end{aligned} \tag{10}$$

where $\lambda_\alpha^2$ are the three eigenvalues of the symmetric tensor $\mathbf{C}$.

*Transversely isotropic model*

Composites are materials composed of a matrix and one or more families of short or long fibers. Fiber-reinforced composites have strong directional properties. Therefore, they can be regarded as anisotropic materials. Since the anisotropic property of the composites appears due to the fibers, the stress at a particular material point depends on the deformation gradient $\mathbf{F}$ and the fiber direction. The direction of a fiber is defined by a unit vector field $\mathbf{a}_0(\mathbf{X})$, $|\mathbf{a}_0| = 1$. Hence, by introducing the tensor product $\mathbf{a}_0 \otimes \mathbf{a}_0$, the following additional invariants can be derived [42]:

$$\mathrm{I}_4(\mathbf{C}, \mathbf{a}_0) = \mathbf{a}_0 \cdot \mathbf{C}\mathbf{a}_0 = \lambda^2 \quad , \qquad \mathrm{I}_5(\mathbf{C}, \mathbf{a}_0) = \mathbf{a}_0 \cdot \mathbf{C}^2\mathbf{a}_0 \quad , \tag{11}$$

and for another additional family of fiber, we obtain:

$$\mathrm{I}_6(\mathbf{C}, \mathbf{g}_0) = \mathbf{g}_0 \cdot \mathbf{C}\mathbf{g}_0 \quad , \qquad \mathrm{I}_7(\mathbf{C}, \mathbf{g}_0) = \mathbf{g}_0 \cdot \mathbf{C}^2\mathbf{g}_0 \quad , \tag{12}$$

Here, $\mathbf{g}_0$ is the vector field describing the orientation of the second fiber family. To account for the angle between the two families of fibers, the following invariant is added:

$$\mathrm{I}_8(\mathbf{C}, \mathbf{a}_0, \mathbf{g}_0) = (\mathbf{a}_0 \cdot \mathbf{g}_0)\mathbf{a}_0 \cdot \mathbf{C}\mathbf{g}_0, \tag{13}$$

where the dot product $\mathbf{a}_0 \cdot \mathbf{g}_0$ determines the cosine of the angle between the two fiber directions.

## 3. Thermodynamically consistent deep-learning model under ambient conditions

In the following section, a PIDL model that satisfies the thermodynamic consistency for dissipative and path-dependent composite materials is proposed.

### *3.1. Feed-forward neural networks*

In deep learning, dense feed-forward neural networks are the basic building block of deep networks and can represent the nonlinear mapping of from inputs $\mathbf{x}$ to predictions $\mathbf{t}$ with a number of consecutive layers $L$ and trainable parameters $\boldsymbol{\omega}$ as

$$\mathcal{F}_{nn}(\boldsymbol{\omega}): \; \mathbf{x} \; \rightarrow \; \mathbf{t}, \tag{14}$$

where the trainable parameters $\boldsymbol{\omega}$ include weights $\mathbf{W}$ and biases $\mathbf{b}$ of each Layer. For each layer, a linear transformation of the inputs $\mathbf{x}$ is applied before a non-linear activation is enforced to obtain the activation of each layer $\mathbf{a}^l$. This can be expressed as

$$\mathbf{a}^l = \Phi^l \left( \mathbf{W}^l \cdot \mathbf{a}^{l-1} + \mathbf{b}^l \right), \; l = 1, 2 \; ... \; L. \tag{15}$$

Here, $\Phi$ is the nonlinear activation function and the first activation $\mathbf{a}^0$ corresponds to the input vector while the last activation $\mathbf{a}^L$ represents the output of the deep network.

*3.2. Long-short term memory network*

While feed-forward deep networks have demonstrated success in capturing the nonlinear behavior of constitutive models [43–46], they fall short when it comes to predicting sequential data $\mathbf{x}^t$ sorted according to real-time. Hence, we implement a multilayered LSTM deep network consisting of several memory cells and gates for remembering and forgetting information in each sequence. The LSTM cell is responsible for predicting the internal variables since the evolution of the internal variables indirectly replicates the history of the deformation.

The architecture of a single LSTM cell is depicted in Fig. 1. In the illustration, $\mathbf{h}$ and $\mathbf{c}$ represent the hidden and cell states at timestep $t$ and $t+1$, respectively. The hidden state encodes the most recent timestep and can be processed at any point to obtain meaningful data. Since the hidden state saves the history information, it is utilized in our PIDL model as described in the next subsection. In contrast, the cell state acts as a global memory of the LSTM network over all timesteps, allowing the LSTM cell to have information on the history of each sequence. The learnable parameters $\boldsymbol{\omega}$ of each component presented in Fig. 1 of the LSTM cell are the input weights $\mathbf{W}$, the recurrent weights $\mathbf{R}$, and the bias $\mathbf{b}$. These matrices are concatenations of the input weights, the recurrent weights, and the bias of each gate as follows:

$$\mathbf{W} = \begin{bmatrix} \mathbf{W}_i \\ \mathbf{W}_f \\ \mathbf{W}_g \\ \mathbf{W}_o \end{bmatrix}, \; \mathbf{R} = \begin{bmatrix} \mathbf{R}_i \\ \mathbf{R}_f \\ \mathbf{R}_g \\ \mathbf{R}_o \end{bmatrix}, \; \mathbf{b} = \begin{bmatrix} \mathbf{b}_i \\ \mathbf{b}_f \\ \mathbf{b}_g \\ \mathbf{b}_o \end{bmatrix}, \tag{16}$$

where $i, f, g$ and $o$ denote the input gate, forget gate, cell candidate, and output gate, respectively, as presented in Fig. 1.

The cell state $\boldsymbol{c}^{t+1}$ at timestep $t+1$ is given by

$$\mathbf{c}^{t+1} = \mathbf{f}^{t+1} \odot \mathbf{c}^t + \mathbf{i}^{t+1} \odot \mathbf{g}^{t+1}. \tag{17}$$

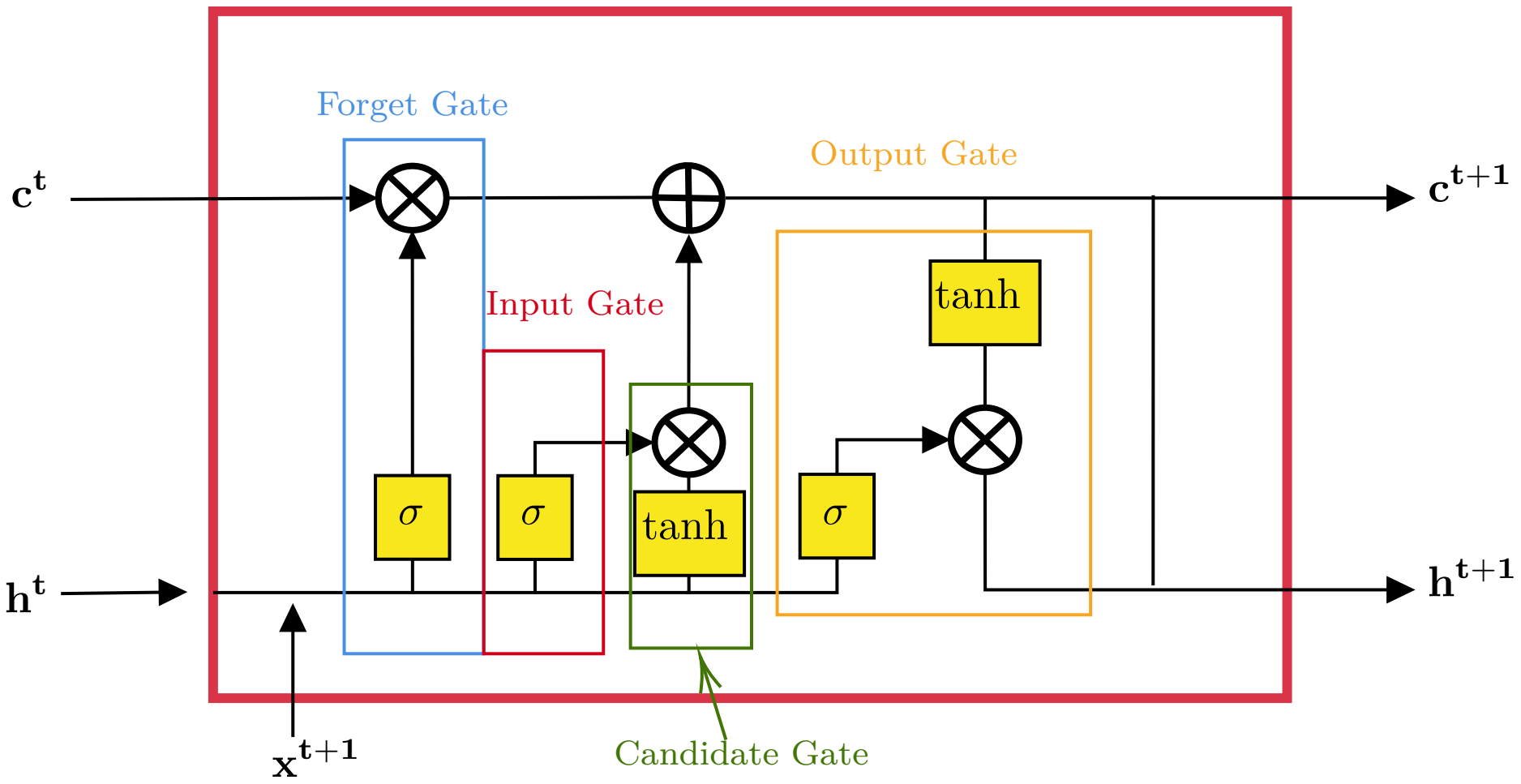

**Figure 1:** A single LSTM cell comprises multiple connected layers. Here, $\sigma$ represents the sigmoid function, while tanh denotes the hyperbolic tangent function. The forget gate, input gate, and output gate regulate the flow of data into the cell. The candidate gate proposes a potential candidate for the cell state.

The operator $\odot$ corresponds to Hadamard product or element-wise product. The hidden state $\mathbf{h}^{t+1}$ can be evaluated as follows:

$$\mathbf{h}^{t+1} = \mathbf{o}^{t+1} \odot \tanh\left(\mathbf{c}^{t+1}\right), \tag{18}$$

where the components of each gate are described by the following equations:

$$\mathbf{i}^{t+1} = \sigma\left(\mathbf{W}_i \mathbf{x}^{\mathbf{t+1}} + \mathbf{R}_i \mathbf{h}^t + \mathbf{b}_i\right), \tag{19}$$

$$\mathbf{f}^{t+1} = \sigma\left(\mathbf{W}_f \mathbf{x}^{t+1} + \mathbf{R}_f \mathbf{h}^t + \mathbf{b}_f\right), \tag{20}$$

$$\mathbf{g}^{t+1} = \tanh\left(\mathbf{W}_g \mathbf{x}^{t+1} + \mathbf{R}_g \mathbf{h}^t + \mathbf{b}_g\right), \tag{21}$$

$$\mathbf{o}^{t+1} = \sigma\left(\mathbf{W}_o \mathbf{x}^{t+1} + \mathbf{R}_o \mathbf{h}^t + \mathbf{b}_o\right). \tag{22}$$

In the equations above, $\mathbf{i}^{t+1}$, $\mathbf{f}^{t+1}$, $\mathbf{g}^{t+1}$ and $\mathbf{o}^{t+1}$ are respectively the components of the input gate, forget gate, cell candidate and output gate, and $\boldsymbol{\sigma}$ represents the sigmoid function. The element-wise product allows each gate to control the data flow into the cell state $\mathbf{c}^{t+1}$ by considering the history of the sequence.

### *3.3. Input*

The goal of the PIDL model is to predict the free-energy function as presented in Eq. (8) under ambient conditions. Therefore, the input vector for our custom PIDL model includes the following values:

1. The kinematic measure of strain $\mathbf{C}^{t+1}$, which describes the deformation path and serves as the first input.
2. The timestep $\Delta \mathrm{t}^t$ to ensure strain-rate effects [24].
3. The moisture content $w_w^{t+1}$.
4. The nanoparticle volume fraction $v_{np}^{t+1}$.
5. The volume fraction of the fibers $v_f^{t+1}$.
6. The temperature $\Theta^{t+1}$.

While the primary input is the tensor $\mathbf{C}^{t+1}$, the invariants are calculated in a second step before using them as an input for the free-energy prediction.

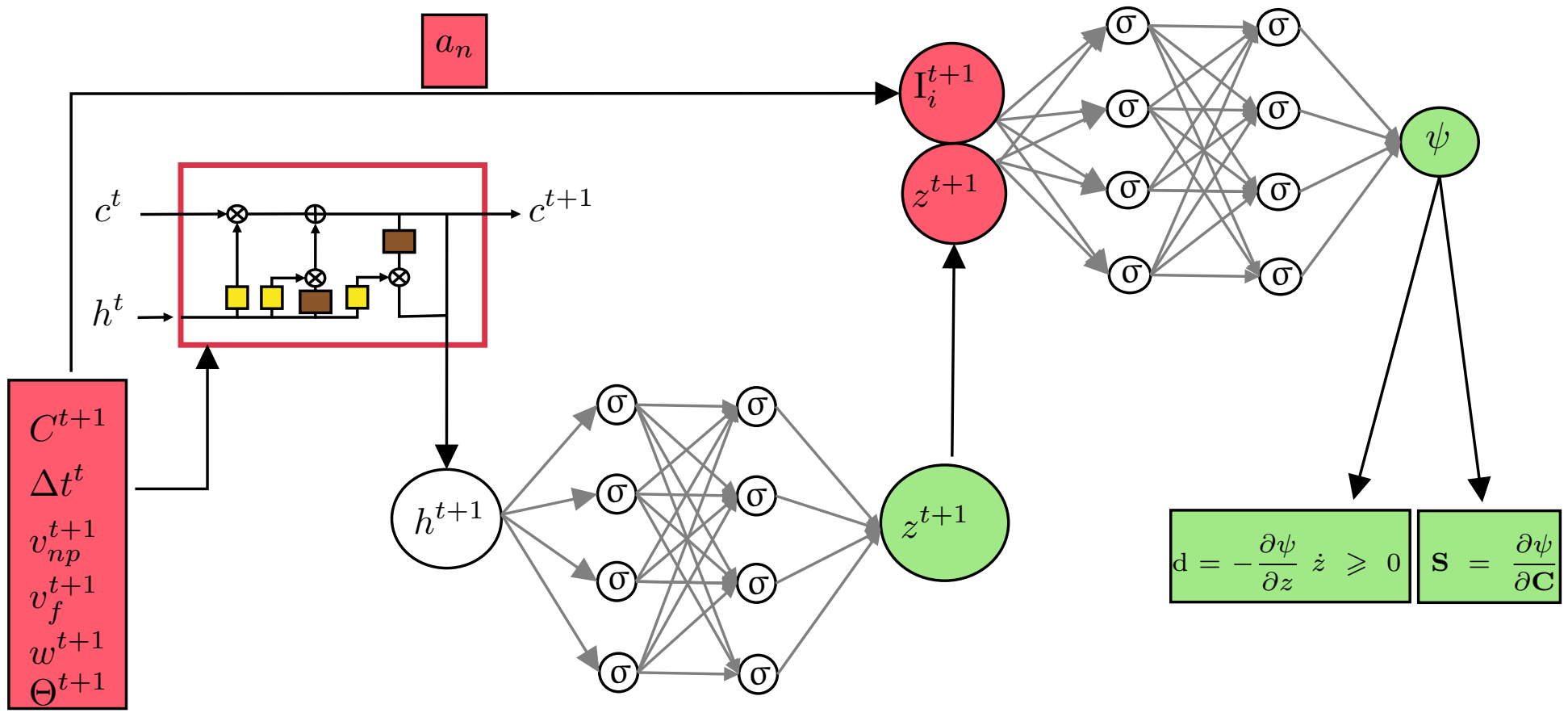

**Figure 2:** An overview of the PIDL model. The red values are input data for the neural networks, and the green data are the outputs of the neural networks. The LSTM cell is connected to another feed-forward neural network. Here, internal variables serve as an input with the invariants to another feed-forward neural network. This leads to the free-energy function, which derives the secondary outputs $\mathbf{S}$ and d.

### *3.4. Architecture of the deep-learning model*

The overall architecture of the model is presented in Fig. 2 and the step-by-step description follows:

1. Initially, the presented input above serves as an input for the LSTM layers, which are responsible to predict the history of the deformation path as:

$$\mathcal{F}_{lstm}(\boldsymbol{\omega}): \ x^{t+1} \ \rightarrow \ h^{t+1}. \tag{23}$$

The initial history of the LSTM layer is initialized with $h^0 = 0$ and $x^{t+1} \in \{\mathbf{C}^{t+1}, \Delta \mathrm{t}^t, w_w^{t+1}, v_{np}^{t+1}, v_f^{t+1}, \Theta^{t+1}\}$.

2. The obtained history output $h^{t+1}$ serves as an input to another feed-forward neural network, whose output leads to the internal variables of the PIDL model:

$$\mathcal{F}_{znn}(\boldsymbol{\omega}): \; [h^{t+1}] \; \rightarrow z^{t+1}. \tag{24}$$

For this particular neural network, the swish activation function and a linear activation function for the last layer are applied.

3. In the next step, the invariants are calculated using the right Cauchy-Green deformation tensor as presented above. These, together with the predicted internal variables, are used as an input to the last feed-forward neural network, which predicts the final free-energy function:

$$\mathcal{F}_{\psi nn}(\boldsymbol{\omega}): \; [z^{t+1}, \mathrm{I}_1(\mathbf{C}), ..., \mathrm{I}_m(\mathbf{C}] \; \rightarrow \; \psi^{t+1}. \tag{25}$$

The free-energy value is then used to compute the second Piola-Kirchhoff stress as presented in Eq. (9) and the dissipation rate from Eq. (7) by using the automatic differentiation. The derivatives of the invariants with respect to the right Cauchy-Green deformation tensor are presented in the appendix A.

4. To calculate the rate of the internal variables $\dot{z}$, we use the following approximation:

$$\dot{z} = \frac{\Delta z}{\Delta t}, \tag{26}$$

where $\Delta z = z^{t+1} - z^t$.

For efficient training, input data are normalized between -1 and 1 by using the following equations:

$$\tilde{x} = \frac{x - m_f}{s_f}, \tag{27}$$

where $m_f = \frac{1}{2}(f_{\max} + f_{\min})$ and $s_f = \frac{1}{2}(f_{\max} - f_{\min})$. Here, $\tilde{x}$ is the scaled value, $f_{\max}$ is the maximum value and $f_{\min}$ is the minimum value of the data.

As described above, the free energy is defined based on the invariants of the right Cauchy-Green tensor and internal variables. The second Piola-Kirchhoff stress and the dissipation rate are computed by automatic differentiation. To ensure normalization of the free energy, the free energy is constrained to $\psi(\mathbf{C} = \mathbf{I}) = 0$. Also, the non-negativeness of the free energy is constrained (i.e., $\psi \; \geqslant \; 0$) by choosing the softplus activation function $\sigma_l(x) = \ln(1 + \exp(x))$ and constrain the weights. Due to the softplus function being $C^{\infty} -$ continuous, we ensure $C^{\infty}$ − continuous functions for the free-energy, stress tensor and the elasticity tensor, which is numerically relevant for implementing the PIDL model within a finite element framework. Also, a linear activation function $\sigma_L(x) = x$ is applied to enforce non-negative weights and biases ($W_L, b_L \geq 0$) in the last layer. This yields a non-negative free-energy function.

*3.5. Model training*

The loss function of the constitutive model is expressed as

$$\mathcal{L} = |\overline{\boldsymbol{\sigma}} - \overline{\boldsymbol{\sigma}}_{pred}| + \beta \mathrm{D}^{\geqslant 0}, \tag{28}$$

where $\beta$ is a weighting parameter and the Cauchy-Stress can be obtained by $\boldsymbol{\sigma} = J^{-1}\, \mathbf{F} \cdot \mathbf{S} \cdot \mathbf{F}^T$ . Although, a constant weighting parameter has shown to be sufficient to train physics-informed neural networks, we enhance the training process and introduce an adaptive rule for the weighting parameter similar to the one presented in [47] by tracking the first- and second-order moments of the gradients during training as follows:

$$\hat{\beta} = \frac{\max_{\theta_n}\{|\nabla_\theta \mathcal{L}_\sigma(\theta_n)|\}}{|\nabla_\theta \beta_i \mathcal{L}_d(\theta_n)|}, \tag{29}$$

where $\theta_n$ are the weights and biases at at step $n$, $\mathcal{L}_\sigma$ is the loss of the stresses, $\mathcal{L}_d$ is the dissipation loss. The parameter $\beta$ is updated as follows:

$$\beta = (1 - \alpha)\beta + \alpha\hat{\beta}. \tag{30}$$

Here, We initialize $\alpha$ to 0.2 and use an exponential decay to reduce the value to 0.05 after 5000 epochs. To ensure $\mathrm{D}^{\geqslant 0}$, the ReLU function is utilized (i.e., ReLU$(x) = max(0, x) \geq 0$), which is positive only if x is positive, therefore leading to $\mathrm{D}^{\geqslant 0}$ = ReLU(-D). Since the PIDL model is supposed to be a constitutive model, the calibration and training of the model is achieved without the information of internal variables or free-energy. Therefore, we do not include these into the loss function and let the PIDL model learn these by itself.

## 4. PIDL model applications

The PIDL model is applied to two cases in the following section. In the first case, the calibration of a classical constitutive viscoelastic-viscoplastic model, including two families of fibers for different moisture, fiber, and nanoparticle content, is performed by utilizing experimental data and subsequently generating synthetic data to train the PIDL model for the three-dimensional case. The accuracy and efficiency of the deep-learning model compared with the classical constitutive model are presented. In the second case, we directly generate data from cyclic loading-unloading experiments conducted on a nanoparticle-filled epoxy system. The experiments include diverse ambient conditions, e.g., temperature, moisture, and nanoparticle volume fraction. Accordingly, the accuracy of the deep-learning model in accurately predicting the material behavior for a material system characterized by a highly nonlinear response with temperature- and moisture dependency is shown. Notably, the deep-learning model solely utilizes experimental data, demonstrating the capability to capture the complex stress-strain response of the material at hand.

### *4.1. Transversely isotropic model with synthetic data for fiber reinforced nanoparticle-filled epoxy nanocomposite*

To rationalize the viscoelastic behavior of fiber-reinforced nanoparticle/epoxy nanocomposites, a viscoelastic-viscoplastic damage model for fiber-reinforced BNP/epoxy nanocomposites is proposed. The stress response is decomposed into an equilibrium part and two viscous parts to capture the nonlinear rate-dependent behavior of the materials. The effect of BNPs and moisture on the stress-strain relationship is considered by defining an amplification factor as a function of the nanofiller and moisture contents. Here, we also take into account the material swelling through moisture.

The total deformation gradient, containing the mechanical deformation, is multiplicatively split into a volumetric and deviatoric part as

$$\mathbf{F} = J^{1/3}\, \mathbf{F}_{iso}, \tag{31}$$

where $J = \det[\mathbf{F}]$ and $\mathbf{F}_{iso}$ are the volumetric deformation and the isochoric deformation gradient, respectively. The volume deformation is further decomposed into two terms: The mechanical compressibility $J_m$ and the moisture-induced swelling $J_w$, leading to an overall volumetric deformation as

$$J = J_m\, J_w, \tag{32}$$

where

$$J_w = 1 + \alpha_w\, w_w. \tag{33}$$

In the equation above, $\alpha_w$ is the moisture swelling coefficient and $w_w$ is the moisture content [7, 48]. The proposed model incorporates experimental characteristics by decomposing the material behavior into a viscoelastic and a viscoplastic part, corresponding to the time-dependent reversible and time-dependent irreversible response, respectively. We further decompose the viscoelastic stress response into a hyperelastic network and a viscous network. The hyperelastic behavior is assumed to originate from the connectivity and stretching of the polymer network and interactions between fiber and nanoparticle reinforcements and polymer chains. In contrast, the viscous network composed of an elastic spring and a viscoelastic dashpot describes the non-equilibrium behavior, taken to be governed by frictional interactions between polymer chains. Additionally, the quasi-irreversible sliding of the molecular chains, resulting in stress softening, also known as the Mullins effect [49, 50], is implemented within the constitutive model. A schematic structure of the model is presented in Fig. 3.

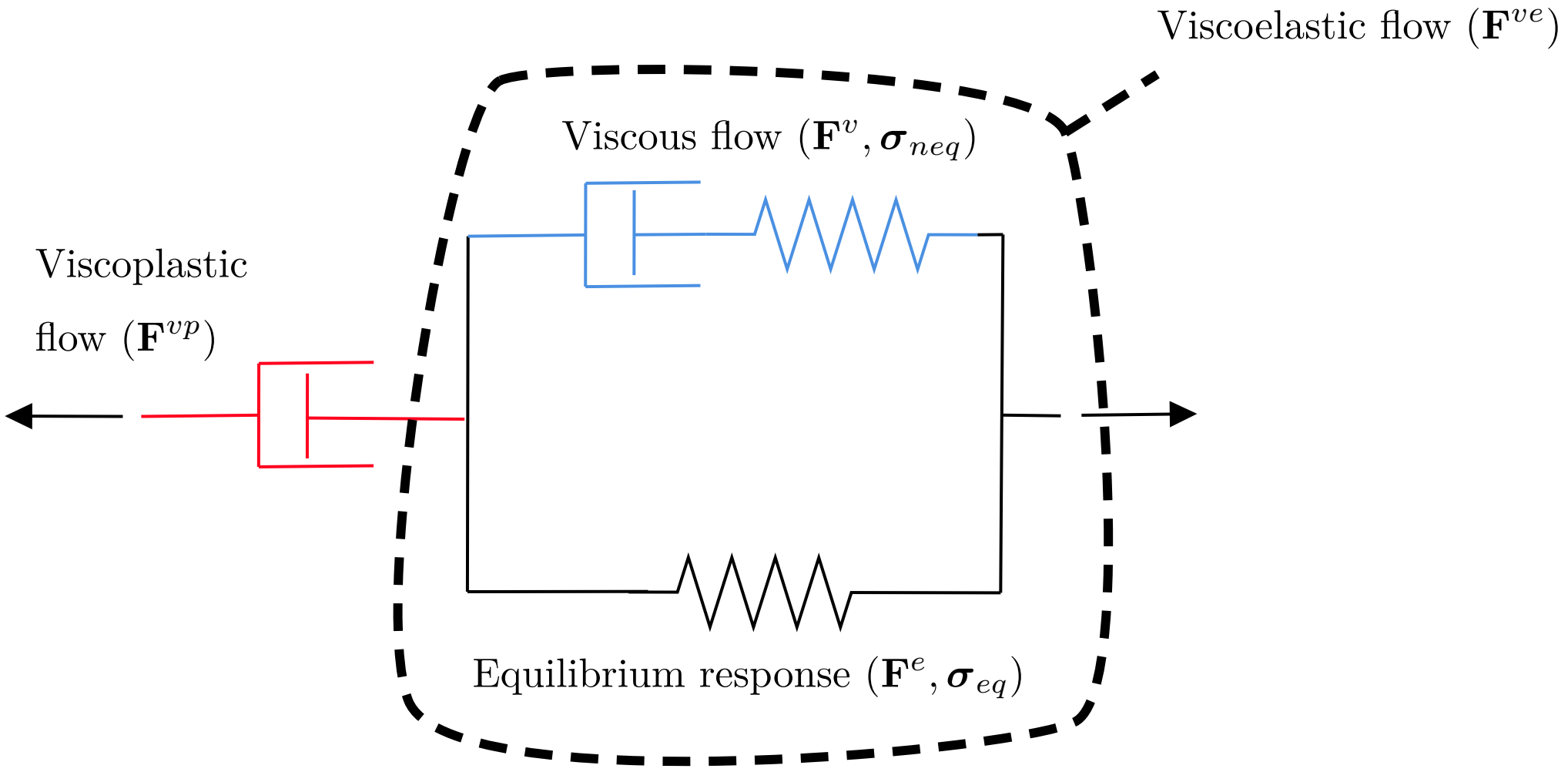

**Figure 3:** One-dimensional schematic of the viscoelastic-viscoplastic constitutive model.

The deviatoric part of the deformation gradient is decomposed into a viscoplastic and a viscoelastic component [51]:

$$\mathbf{F}_{iso} = \mathbf{F}_{iso}^{ve}\mathbf{F}_{iso}^{vp}. \tag{34}$$

Also, the viscoelastic deformation gradient is split into an elastic and an inelastic part as

$$\mathbf{F}_{iso}^{ve} = \mathbf{F}_{iso}^{e}\mathbf{F}_{iso}^{v}. \tag{35}$$

Accordingly, similar decompositions are obtained for the left Cauchy-Green deformation tensors:

$$\mathbf{B}_{iso} = \mathbf{F}_{iso}\ \mathbf{F}_{iso}^{T}, \tag{36}$$

$$\mathbf{B}_{iso}^{v} = \mathbf{F}_{iso}^{v}\ \mathbf{F}_{iso}^{vT}, \tag{37}$$

$$\mathbf{B}_{iso}^{e} = \mathbf{F}_{iso}^{e}\ \mathbf{F}_{iso}^{eT}. \tag{38}$$

The total Cauchy stress in damaged state is given by

$$\boldsymbol{\sigma} = (1 - \mathrm{d})(\boldsymbol{\sigma}_{eq} + \boldsymbol{\sigma}_{neq}), \tag{39}$$

where $\boldsymbol{\sigma}_{eq}$ and $\boldsymbol{\sigma}_{neq}$ are the portions of the total stress originating from the hyperelastic rubbery and the viscoelastic behavior of the material, respectively. The stresses are finally expressed as

$$\begin{aligned}\boldsymbol{\sigma}_{eq} = \frac{2}{J_{ve}} \sum_{i=1}^{N} \Bigg[ \frac{\partial \psi_{eq}^{i}}{\partial \mathrm{I}_{ve_1}} \mathrm{dev} \left[ \mathbf{B}_{iso}^{ve} \right] + \frac{\partial \psi_{eq}^{i} \mathrm{I}_{ve_4}^{i}}{\partial \mathrm{I}_{ve_4}^{i}} \left( \mathbf{a}_e^i \otimes \mathbf{a}_e^i - \frac{1}{3}\mathbf{I} \right] \Bigg) + \\ \frac{\partial \psi_{eq}^{i}}{\partial \mathrm{I}_{ve5}^{i}} \left( \mathrm{I}_{ve_4}^{i} \left( \mathbf{a}_e^i \otimes \mathbf{B}_{iso}^{ve} \cdot \mathbf{a}_e^i + \mathbf{a}_e^i \cdot \mathbf{B}_{iso}^{ve} \otimes \mathbf{a}_e^i \right) - \frac{2}{3} \mathrm{I}_{ve5}^{i} \mathbf{I} \right) \Bigg],\end{aligned} \tag{40}$$

and

$$\begin{aligned}\boldsymbol{\sigma}_{neq} = \frac{2}{J_{e}} \sum_{i=1}^{N} \Bigg[ \frac{\partial \psi_{neq}^{i}}{\partial \mathrm{I}_{e_1}} \mathrm{dev} \left[ \mathbf{B}_{iso}^{e} \right] + \frac{\partial \psi_{neq}^{i} \mathrm{I}_{e_4}^{i}}{\partial \mathrm{I}_{e_4}^{i}} \left( \mathbf{a}_v^i \otimes \mathbf{a}_v^i - \frac{1}{3}\mathbf{I} \right] \Bigg) + \\ \frac{\partial \psi_{neq}^{i}}{\partial \mathrm{I}_{e5}^{i}} \left( \mathrm{I}_{e_4}^{i} \left( \mathbf{a}_v^i \otimes \mathbf{B}_{iso}^{e} \cdot \mathbf{a}_v^i + \mathbf{a}_v^i \cdot \mathbf{B}_{iso}^{e} \otimes \mathbf{a}_v^i \right) - \frac{2}{3} \mathrm{I}_{e5}^{i} \mathbf{I} \right) \Bigg] + \frac{\partial \psi_{neq}}{\partial J_e} \mathbf{I},\end{aligned} \tag{41}$$

where the superscript $i$ denotes the $i$th family of fibers, $\mathbf{I}$ is the second order unit tensor. $\mathbf{a}_e^i = \frac{\mathbf{F_{ve} a_0^i}}{\sqrt{\mathrm{I}_{ve_4}^i}}$,$\mathbf{a}_v^i = \frac{\mathbf{F_e a_0^i}}{\sqrt{\mathrm{I}_{e_4}^i}}$ are the current fiber direction for each family of fibers and $\frac{\partial \psi^i}{\partial \mathrm{I}_j}$ $(j =$

1, 4 and 5) are calculated as follows [52]:

$$\begin{aligned}
\frac{\partial\psi^i}{\partial \mathrm{I}_1}\left(\mathrm{I}_4^i,\mu_m\right) &= \frac{1}{2}g\left[f\left(\mathrm{I}_4^i\right),\nu_f,0.4\right]\mu_m,\\
\frac{\partial\psi^i}{\partial \mathrm{I}_4}\left(\mathrm{I}_1,\mathrm{I}_4^i,I_5^5,\mu_m\right) &= \frac{1}{2}\mu_m[\nu_f\frac{\partial f\left(\mathrm{I}_4^i\right)}{\partial \mathrm{I}_4^i}\left(\mathrm{I}_4^i\ +\ 2\mathrm{I}_4^{i^{-1/2}}-3\right)+\left(\nu_m+\nu_f f\left(\mathrm{I}_4^i\right)\right)\left(1-\mathrm{I}_4^{i^{-3/2}}\right)\\
&\quad -g\left[f\left(\mathrm{I}_4^i\right),\nu_f,1\right]\left(\mathrm{I}_5^i\mathrm{I}_4^{i^{-2}}+1\right)\ +\ g\left[f\left(\mathrm{I}_4^i\right),\nu_f,0.4\right]\left(\mathrm{I}_5^i\mathrm{I}_4^{i^{-2}}+\mathrm{I}_4^{i^{-3/2}}\right)\\
&\quad +\frac{\mathrm{I}_5^i-\mathrm{I}_4^{i^2}}{2\mathrm{I}_4^i}\frac{\partial g\left[f\left(\mathrm{I}_4^i\right),\nu_f,0.4\right]}{\partial \mathrm{I}_4^i}+\frac{1}{2}\left(\mathrm{I}_1-\frac{I_5^i+2\sqrt{\mathrm{I}_4^i}}{I_4^i}\right)\frac{\partial g\left(f\left(\mathrm{I}_4^i\right),\nu_f,0.4\right]}{\partial \mathrm{I}_4^i}],\\
\frac{\partial\psi^i}{\partial \mathrm{I}_5}\left(\mathrm{I}_4^i,\mu_m\right) &= \frac{1}{2\mathrm{I}_4^i}\left(g\left[f\left(\mathrm{I}_4^i\right),\nu_f,1\right]\ -\ g\left[f\left(\mathrm{I}_4^i\right),\nu_f,0.4\right]\right)\mu_m\\
\frac{\partial\psi_{neq}}{\partial J_e} &= k_\nu\left(J_m-\frac{1}{J_m}\right).
\end{aligned} \tag{42}$$

In above equations $\mathrm{I}_1=\mathrm{I}_{ve1},\mathrm{I}_4=\mathrm{I}_{ve4}$ and $\mathrm{I}_5=\mathrm{I}_{ve5}$ for the equilibrium response $\boldsymbol{\sigma}_{neq}$ and $\mathrm{I}_1=\mathrm{I}_{e1},\mathrm{I}_4=\mathrm{I}_{e4}$ and $\mathrm{I}_5=\mathrm{I}_{e5}$ for the non-equilibrium response $\boldsymbol{\sigma}_{eq}$. $\nu_m$ and $\nu_f$ are volume fractions of matrix and fibers, $f(\mathrm{I}_4)=a_1+a_2\exp\left[a_3(\mathrm{I}_4-1)\right]$, describes the stiffness ratio between the fiber and the matrix [53], where $a_1,a_2$ and $a_3$ are positive parameters and $g[f(\mathrm{I}_4),\nu_f,\zeta]$ is the ratio of effective shear moduli between composite $\mu_C$ and matrix $\mu_m$ shear moduli and is obtained as follows:

$$g[f(\mathrm{I}_4),\nu_f,\zeta]=\frac{\mu_C}{\mu_m}=\frac{(1+\zeta\nu_F)f(\mathrm{I}_4)+(1-\nu_F)\zeta}{(1-\nu_F)f(\mathrm{I}_4)+(\zeta+\nu_F)}, \tag{43}$$

where $\zeta$ considers in-plane shear and transverse shear as follows:

$$\zeta=\begin{cases}1 & \mathrm{in-plane\ shear}\\ 0.4 & \mathrm{transverse\ shear}\end{cases}. \tag{44}$$

Here, a modified Guth-Gold model [54] is used to obtain the effective stiffness of the nanoparticle-modified epoxy system. Also, adding a moisture dependency to the effective stiffness results in the following amplification factor [7]

$$\mathrm{X}=\ \left(1\ +\ 5v_{np}\ +\ 18v_{np}^2\right)\left(1+\ 0.057w_w^2\ -\ 9.5w_w\right), \tag{45}$$

where $v_{np}$ is the volume fraction of BNPs and $w_w$ represents the moisture content.

The total velocity gradient of the viscoelastic network, $\mathbf{L}^{ve}=\ \dot{\mathbf{F}}^{ve}\left(\mathbf{F}^{ve}\right)^{-1}$ , can be decomposed into an elastic and a viscous component analogously to Eq. (35)

$$\mathbf{L}^{ve}=\ \mathbf{L}^e\ +\ \mathbf{F}^e\mathbf{L}^e\mathbf{F}^{e-1}\ =\ \mathbf{L}^e\ +\ \tilde{\mathbf{L}}^v, \tag{46}$$

and

$$\tilde{\mathbf{L}}^{v} = \dot{\mathbf{F}}^{v}\mathbf{F}^{v-1} = \tilde{\mathbf{D}}^{v} + \tilde{\mathbf{W}}^{v}. \tag{47}$$

Here, $\tilde{\mathbf{D}}^{v}$ represents the rate of the viscous deformation and $\mathbf{W}^{v}$ is a skew-symmetric tensor representing the rate of stretching and spin, respectively. The intermediate state can be made unique by prescribing $\tilde{\mathbf{W}}^{v} = 0$. The rate of the viscoelastic flow can be described by

$$\tilde{\mathbf{D}}^{v} = \frac{\dot{\varepsilon}^{v}}{\tau_{neq}} \operatorname{dev}\left[\boldsymbol{\sigma}_{neq}^{'}\right] \tag{48}$$

where $\tau_{neq} = \| \operatorname{dev}[\boldsymbol{\sigma}_{neq}] \|_F$ represents the Frobenius norm of the driving stress, $\dot{\varepsilon}^{v}$ is the viscous flow and $\boldsymbol{\sigma}_{neq}^{'} = \mathbf{R}_e^T\boldsymbol{\sigma}_{neq}\mathbf{R}_e$ represents the stress acting on the viscous component in its relaxed configuration. The viscous flow is defined by the Argon model:

$$\dot{\varepsilon}_{v} = \dot{\varepsilon}_0 \exp\left[\frac{\Delta \mathrm{H}}{k_b \mathrm{T}}\left(\left(\frac{\tau_{neq}}{\tau_0}\right)^{m} - 1\right)\right], \tag{49}$$

where $k_b$, $\dot{\varepsilon}_0$, $\Delta$H and $\tau_0$ are the Boltzmann constant, a pre-exponential factor, the activation energy and the athermal yield stress. Here, a sigmoid function is used for the athermal yield stress modification as follows:

$$\tau_0 = \mathrm{y}_0 + \frac{\mathrm{a}_s}{1 + \exp\left(-\frac{\left(\Lambda_{chain}^{max} - \mathrm{x}_0\right)}{\mathrm{b}_s}\right)}. \tag{50}$$

In summary, the time derivative of $\dot{\mathbf{F}}^{v}$ can be derived from Eq. (48) and Eq. (47) as follows:

$$\dot{\mathbf{F}}^{v} = \mathbf{F}^{e-1}\dot{\varepsilon}^{v}\frac{\operatorname{dev}\left[\boldsymbol{\sigma}_{neq}^{'}\right]}{\tau_{neq}}\mathbf{F}^{ve}. \tag{51}$$

The viscoplastic velocity gradient is considered to be additively decomposed into the symmetric rate of stretching and the skew-symmetric rate of spinning:

$$\tilde{\mathbf{L}}^{vp} = \dot{\mathbf{F}}^{vp}\mathbf{F}^{vp-1} = \tilde{\mathbf{D}}^{vp} + \tilde{\mathbf{W}}^{vp}, \tag{52}$$

and $\tilde{\mathbf{W}}^{vp} = 0$, leading to:

$$\tilde{\mathbf{D}}^{vp} = \frac{\dot{\varepsilon}^{vp}}{\tau_{tot}} \operatorname{dev}\left[\boldsymbol{\sigma}_{tot}^{'}\right], \tag{53}$$

where $\operatorname{dev}\left[\boldsymbol{\sigma}_{tot}^{'}\right]$ is the total deviatoric stress in its relaxed configuration and $\tau_{tot}$ is the Frobenius norm of the total stress. To characterize the viscoplastic flow $\dot{\varepsilon}^{vp}$, a simple phenomenological representation is implemented similar to [55] as follows:

$$\dot{\varepsilon}^{vp} = \begin{cases} 0 & \tau_{tot} < \sigma_0 \\ \mathrm{a}(\epsilon - \epsilon_0)^{\mathrm{b}}\dot{\epsilon} & \tau_{tot} \geq \sigma_0 \end{cases}, \tag{54}$$

Here, $a$, $b$ and $\sigma_0$ are material parameters. $\epsilon_0$ is the stress at which the viscoplastic flow is activated. This is represented by the Frobenius norm of the Green strain tensor $\| \mathbf{E} \|_F$, which is derived from the deformation gradient

$$\mathbf{E} = \frac{1}{2} (\mathbf{F}^T \mathbf{F} - \mathbf{I}), \tag{55}$$

and $\dot{\epsilon}$ is the strain rate of the effective strain $\| \mathbf{E} \|_F$, thus introducing a simple strain-rate dependency of the viscoplastic flow. Analogous to Eq. (51), the time derivative of the viscoplastic deformation gradient is given by

$$\dot{\mathbf{F}}^{vp} = \mathbf{F}^{ve-1} \dot{\varepsilon}^{vp} \frac{\mathrm{dev}\left[\boldsymbol{\sigma}'_{tot}\right]}{\tau_{tot}} \mathbf{F}_{iso}, \tag{56}$$

characterizing the rate kinematics of the viscoplastic flow. The viscous and viscoplastic deformation gradients are obtained at the end of a time increment using the Euler backward time integration.

d $\in [0, 1)$ is the scalar damage variable and its evolution obeys

$$\dot{\mathrm{d}} = \mathrm{A}(1 - \mathrm{d})\dot{\Lambda}^{max}_{chain}, \tag{57}$$

where A is a material parameter calibrated using experimental data, $\Lambda_{chain} = \sqrt{\mathrm{tr}[\mathbf{B}_{iso}]/3}$, $\dot{\Lambda}^{max}_{chain}$ takes the following form:

$$\dot{\Lambda}^{max}_{chain} = \begin{cases} 0 & \Lambda_{chain} < \Lambda^{max}_{chain} \\ \dot{\Lambda}_{chain} & \Lambda_{chain} \geq \Lambda^{max}_{chain} \end{cases} \tag{58}$$

For a detailed discussion of the viscoelastic-viscoplastic constitutive model, the reader is referred to [24].

### *4.2. Numerical results*

In the following, the calibration of the classical transversely isotropic model for the fiber reinforced nanoparticle-filled epoxy nanocomposite with moisture content using experimental results is shown. Accordingly, a data generation scheme using the classical constitutive model is derived and the PIDL model validation is presented.

#### *Constitutive model calibration*

Concerning the tensile tests for the calibration, specimens with a thickness of 2.2 mm are produced and tested according to testing standard DIN EN ISO 5272, using an extensometer to measure the elongation of samples. The stress-strain curves are recorded with a test speed of 1 mm/min. The specimens were firstly dried for 7 days at 50 °C and another 21 days at 70 °C. Secondly, a set of specimens are conditioned at 70 °C and 85% relative humidity until the saturated state is reached. The overall conditioning time was 115 days.

To calibrate the material parameters, the stress-strain relationships of two types of specimens epoxy plus 5% BNP volume fraction, 25% GF volume fraction and epoxy plus 20% BNP volume fraction, 20% GF volume fraction at dry and saturated state obtained from the constitutive model is compared with those of experimental data. An object function of differences between model results and experimental data is numerically minimized using the single-objective genetic algorithm method. The GFs are in-plane randomly-oriented, which are represented by two families of fibers with the initial fiber directions of $\mathbf{a}_0 = [1 \quad 0 \quad 0]$ and $\mathbf{g}_0 = [0 \quad 1 \quad 0]$. The experimentally identified parameters of the constitutive model are listed in Table 1. While the material parameters regarding the viscoelastic dashpot are derived from the argon theory and identified using atomistic simulations [10, 56], the remaining parameter identification approach is presented in [24]. The prediction of the calibrated classical constitutive model is presented in Fig. 4 for the saturated (blue lines) and dry state (black lines). The agreement between experimental data and constitutive model prediction confirms the predictive capability of the implemented viscoelastic-viscoplastic damage model on the mechanical behavior of the fiber reinforced nanocomposite.

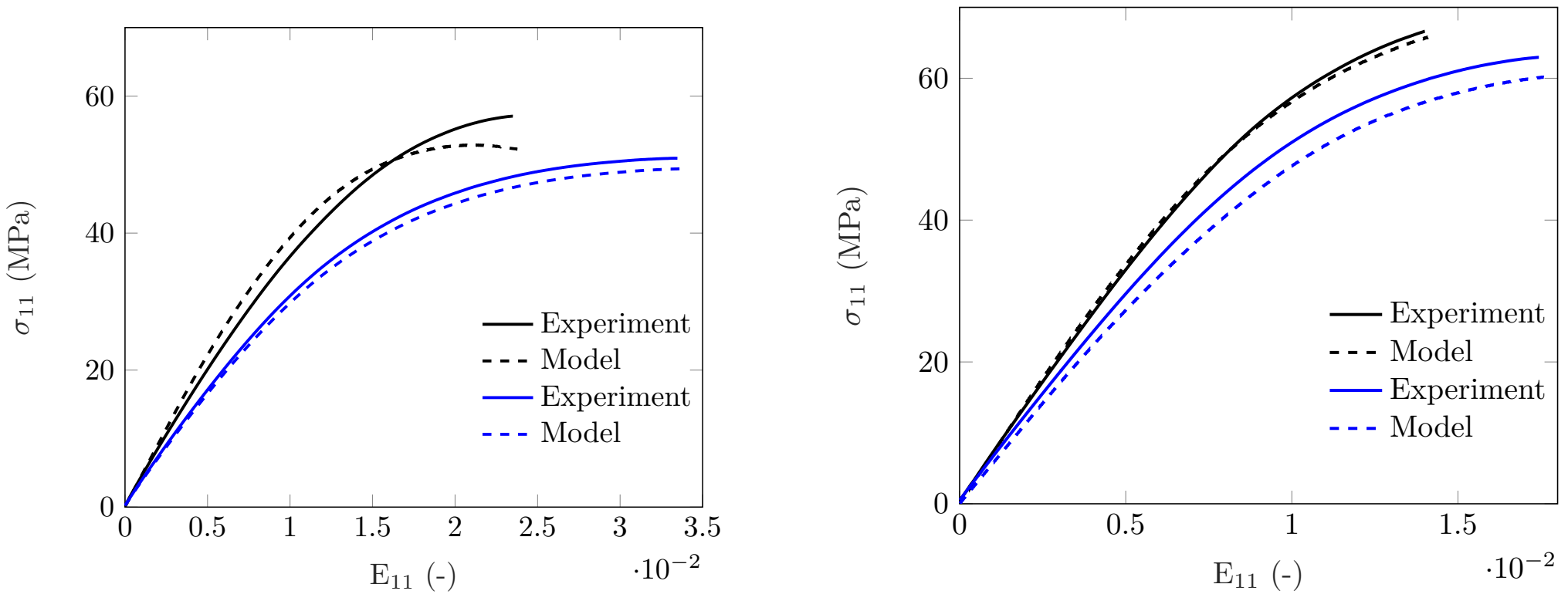

**(a)** 5% BNP/22% GF

**(b)** 20% BNP/20% GF

**Figure 4:** Experimental and predicted stress-strain behavior of the constitutive model (dashed lines) and experiments (solid lines) for dry (black lines) and saturated (blue lines) state.

**Table 1:** Materials parameters of the viscoelastic-viscoplastic damage model

| | Parameter | Value | Equation | References |
|---|---|---|---|---|
| Equilibrium shear modulus | $\mu_m^{eq}$(MPa) | 525 | 42 | |
| Non-equilibrium shear modulus | $\mu_m^{neq}$(MPa) | 295 | 42 | |
| Volumetric bulk modulus | $\kappa_\nu$(MPa) | 1311 | 42 | |
| Viscoelastic dashpot | $\dot{\varepsilon}_0$ ($s^{-1}$) | 1.0447 x $10^{12}$ | 49 | [56] |
| | $\Delta H$(J) | 1.977 x $10^{-19}$ | 49 | [56] |
| | $m$ | 0.837 | 49 | |
| | $y_0$ | 80 | 50 | |
| | $x_0$ | 1.72 | 50 | |
| | $b_s$ | 0.394 | 50 | |
| | $a_s$ | -40.17 | 50 | |
| Viscoplastic dashpot | a | 48.37 | 54 | |
| | b | 1.02 | 54 | |
| | $\sigma_0$(MPa) | 5.5 | 54 | |
| Damage | A | 943.87 | 57 | |
| Moisture swelling coefficient | $a_w$ | 0.039 | 33 | [48] |
| Stiffness ratio parameter | $a_1$ | 9 | 42 | |
| Stiffness ratio parameter | $a_2$ | 1 | 42 | |
| Stiffness ratio parameter | $a_3$ | 1 | 42 | |

*Data generation and training*

We combine the PIDL model with the classical constitutive model above using the stress tensor of the undamaged material. The stress tensor of the damaged material is then calculated using $\boldsymbol{\sigma} = (1 - \mathrm{d})(\boldsymbol{\sigma}_{pred})$ in the postprocessing. Therefore, we attain the presented damage model and use the PIDL model to predict the stress tensor of the undamaged material. It is a hybrid form of combining the classical constitutive model with the deep-learning model.

The driving force behind the generation of loading paths in the finite deformation model is the deformation gradient $\mathbf{F}$. Each component $F_{ij}$ (where $i, j = 1, \ldots, 3$) corresponds to an unique loading scenario. Consequently, synthetic data is generated by varying the components of the deformation gradient in a spatio-temporal space. Initially, constraints are imposed on the components of the deformation gradient as follows:

$$\mathrm{F}_{ij} \in \begin{cases} [0.98 \quad 1.02], & \text{when i } = \text{ j} \\ [-0.02 \quad 0.02], & \text{when i } \neq \text{ j} \end{cases} . \tag{59}$$

To cover the nine-dimensional spatial space with sufficient random points, quasi-random numbers are produced using the Halton sequence generation algorithm [57]. This approach yields uniform samples within the space, resulting in a more effective sampling of the region. As presented in Fig. 5, the resulting points in the spatial space adequately span the bounded domain compared with the pseudorandom algorithm used in most algorithms. Table 2 summarizes the step-by-step algorithm for the generation of training data in the nine-dimensional spatio-temporal space. For a more specific and detailed discussion, the reader is referred to [24].

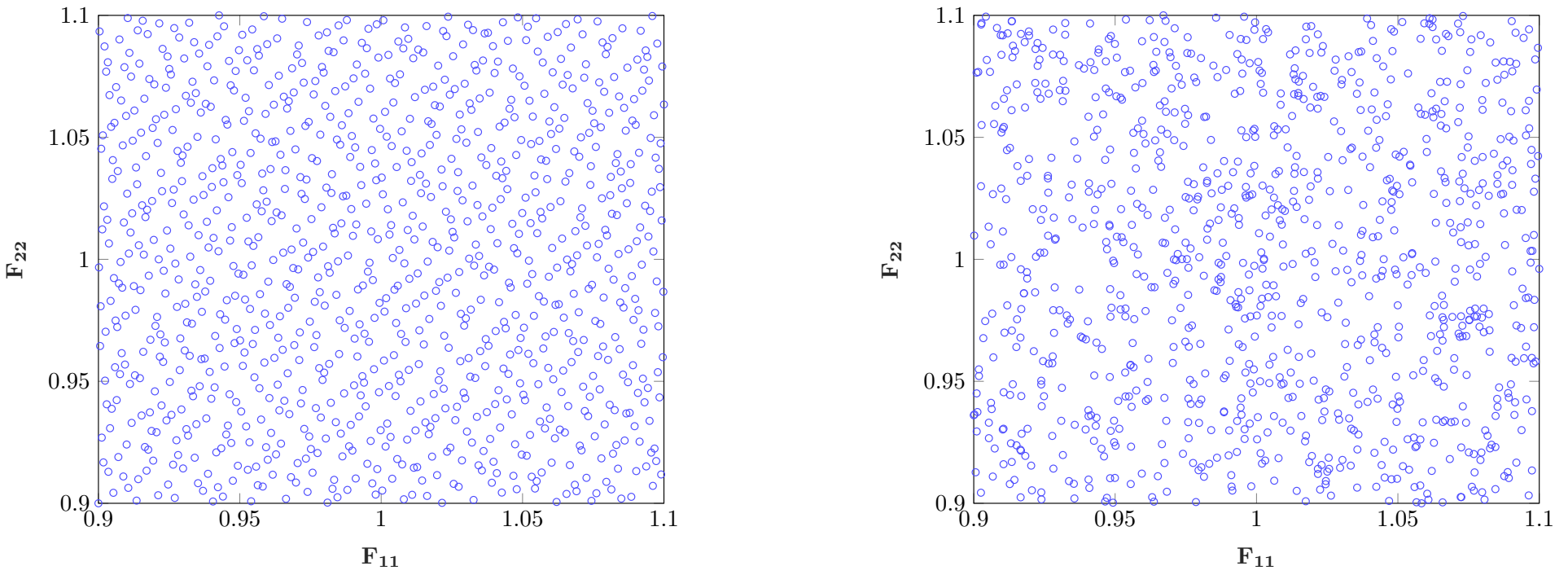

**(a)** Uniform data points  **(b)** Pseudorandom data points

**Figure 5:** Example of the generated data points in a two dimensional space using the Halton sequence algorithm (a) and the pseudorandom MATLAB algorithm (b).

**Table 2:** Summary of the step-by-step algorithm for generation of a single training sequence.

| |
|---|
| 1. Generate uniform data for the nine components. |
| of the deformation gradient in a range defined by 60. |
| 2. Define P as the number of points to be visited. |
| 3. Generate loading path within the 9-dimensional space |
| 4. Calculate a representative strain rate as: $\dot{\epsilon} = \Vert \mathbf{E} \Vert_F / \Delta \mathrm{t}$ , |
| where $\mathbf{E}$ is presented in Eq. (55). |
| 5. If $\dot{\epsilon} > 1 \cdot 10^{-5}$ *and* $\dot{\epsilon} < 1 \cdot 10^{-3}$ GOTO step 6 else GOTO step 3. |
| 6. Integrate the constitutive model and obtain $\boldsymbol{\sigma}_{tot}$. |
| 7. Create the input sequence $\mathbf{x}$ and the output sequence $\boldsymbol{\sigma}_{tot}$ for training. |

A total of 1000 sequences are generated for the training of the PIDL model and another 200 sequences for the validation process. We include into the training process the invariants $\mathrm{I}(\mathbf{C}) \in \{\mathrm{I}_1, \mathrm{I}_2, \mathrm{I}_3, \mathrm{I}_4, \mathrm{I}_5, \mathrm{I}_6, \mathrm{I}_7, \mathrm{I}_8\}$ from Section 2.2 and the hyperparameters of the PIDL model are presented in Table 3.

**Table 3:** Hyperparameters of the PIDL model for the first application case

| Hyperparameter | Value |
|---|---|
| Learning rate $\alpha$ | $1e-3$ |
| Epochs | 5000 |
| Batch size | 32 |
| Hidden layers ($\mathcal{F}_{lstm}, \mathcal{F}_{znn}, \mathcal{F}_{\psi nn}$) | 2 |
| Number of neurons per hidden layer ($\mathcal{F}_{lstm}, \mathcal{F}_{znn}, \mathcal{F}_{\psi nn}$) | 100 |

Firstly, the internal variables are tuned by training the model with 2 to 15 internal variables. The loss in Fig. 6 shows that the optimum internal variables we can consider are 10. Therefore, we continue with 10 internal variables for the remaining results. Also, the validation loss reaches a value of 0.008, which is low enough to validate the results.

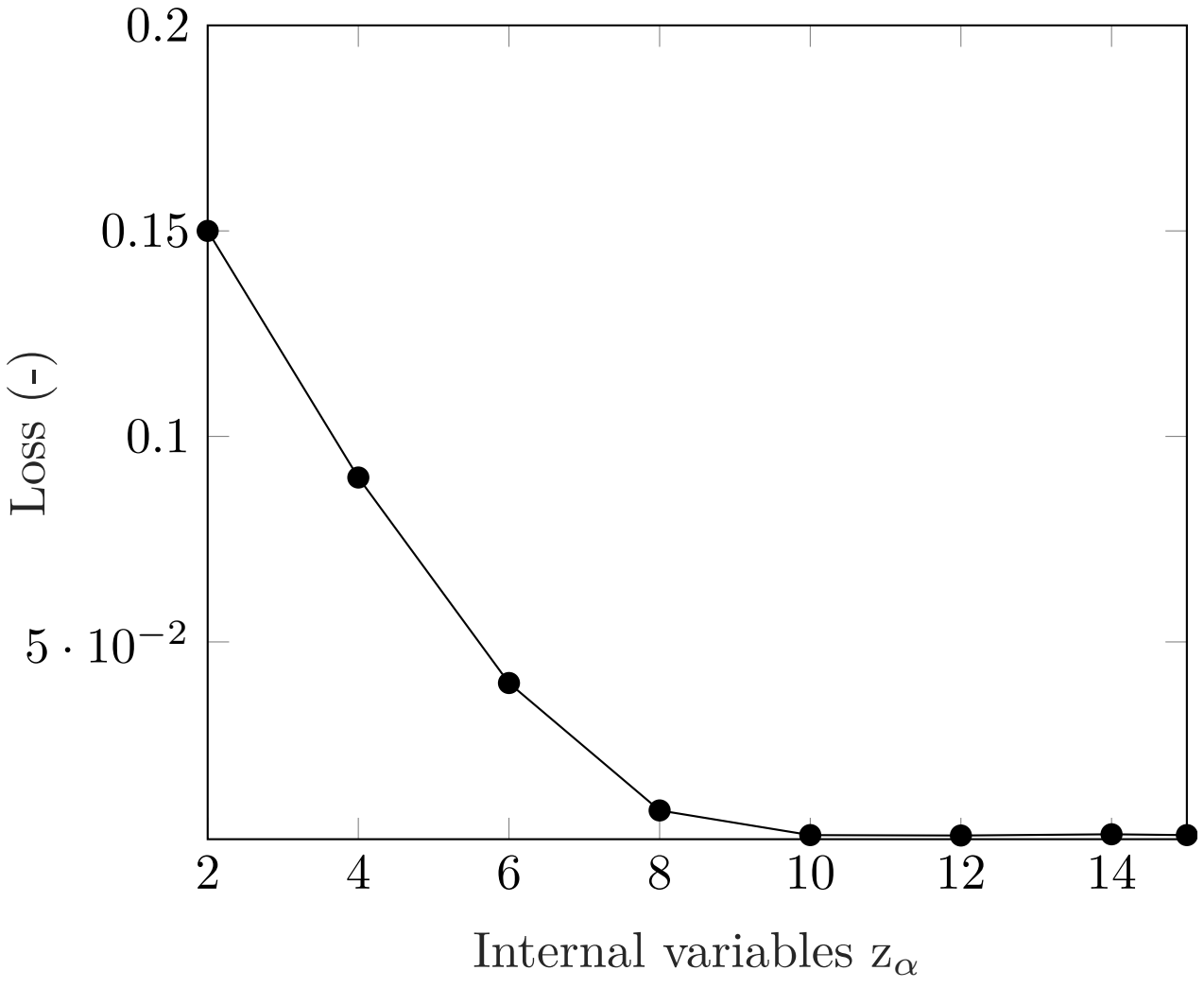

**Figure 6:** Final training loss as a function of internal variables.

*Validation and discussion*

As mentioned above, the validation of the model is done using 200 additional sequences leading to a loss of 0.008. To validate the PIDL model, data with a higher deformation gradient component is generated as follows:

$$\mathrm{F}_{ij} \in \begin{cases} [0.97 \quad 1.03], & \text{when i } = \text{ j} \\ [-0.03 \quad 0.03], & \text{when i } \neq \text{ j} \end{cases}. \tag{60}$$

This ensures a validation for the interpolation and extrapolation of the PIDL model. Fig. 7(a) and (b) illustrate the behavior of the fiber- and nanoparticle-reinforced epoxy under a random three-dimensional loading path, as depicted in Fig. 7(c). The solid lines represent the classical constitutive model and the dashed lines are predicted by the PIDL model. In this case, two data points are visited according to Table 2, leading to a complex loading path for each deformation gradient component. Remarkably, the model accurately predicts the stress-strain behavior under this loading scenario, and extrapolation over a strain of $\mathbf{E} = 0.02$ for the diagonal part is demonstrated. As can be seen, the model can also predict the constraints in different directions due to the Poisson's ratio, resulting in additional stress without a corresponding increase in deformation.

Another validation result is presented in Fig. 8(a) and (b) for the loading-unloading scenario as shown in Fig. 8(c) and a different strain rate in comparison to Fig. 7. A comparison between the experimental and deep learning data reveals that the proposed PIDL model can predict the stress-strain behavior for a rate-dependent material at a dry/saturated state and extrapolate over the trained strain boundaries.

Fig. 9(a) presents the free-energy value as a function of loading steps for the 3D case of Fig. 8. We observe a convex behavior of the free energy, which is crucial to enforcing the Clausius-Duhem inequality and calculating the material tangent essential for finite element implementation. The minimum of the free energy is at the initial state, which was enforced within our deep learning formulation in Section 3.4. We also observe maximum free energy after approximately 520 loading steps, correlating with the stress maxima at the loading peak shown in Fig. 8(c). As expected, the free energy of the stiffer model (20%BNP/20%GF) reaches higher values, and during the unloading path (post 520 loading steps), a convex decrease of the free energy is revealed. Notably, a kink in the free-energy function at approximately 95 loading steps aligns with the deformation path in Fig. 8, where a kink of the deformation is observed at the same loading step. Fig. 9(b) shows the dissipation of the example, where we observe a highly nonlinear behavior. Noteworthy, the dissipation reaches two peaks at the inflection points of the free energy and decreases towards the end of the deformation path. The peaks also correlate with the evolution of the internal variables depicted in Fig. 9(c), where higher movement is observed at the beginning of the deformation path up to the second dissipation maximum and more linear behavior can be seen after the turning point of the deformation at the 520 loading step mark.

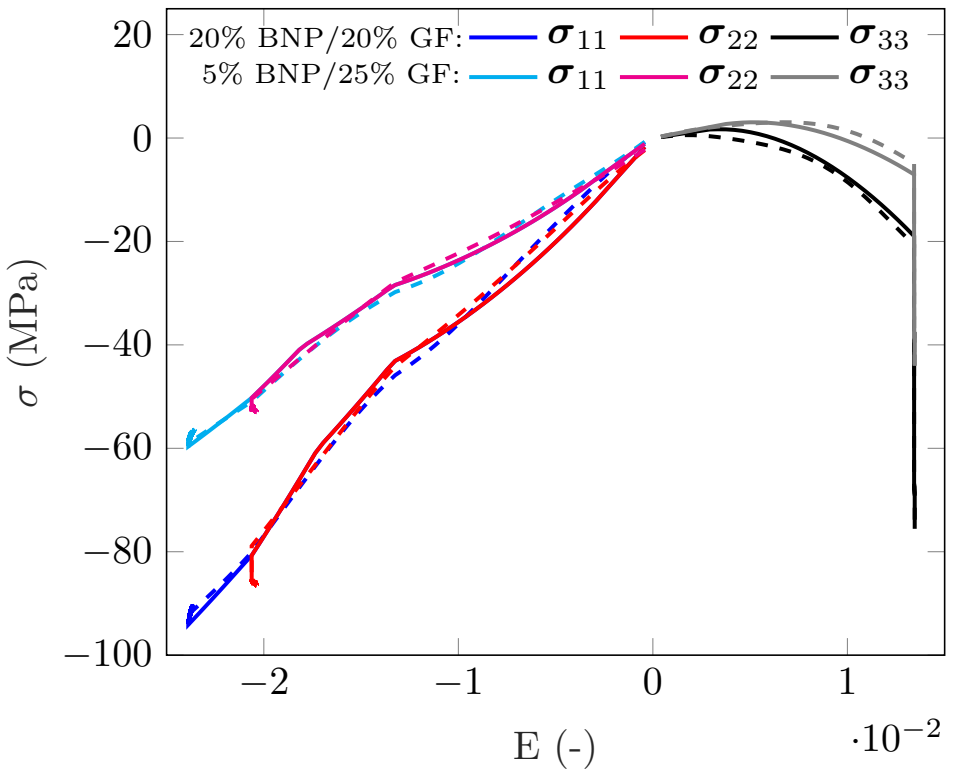

**(a)**

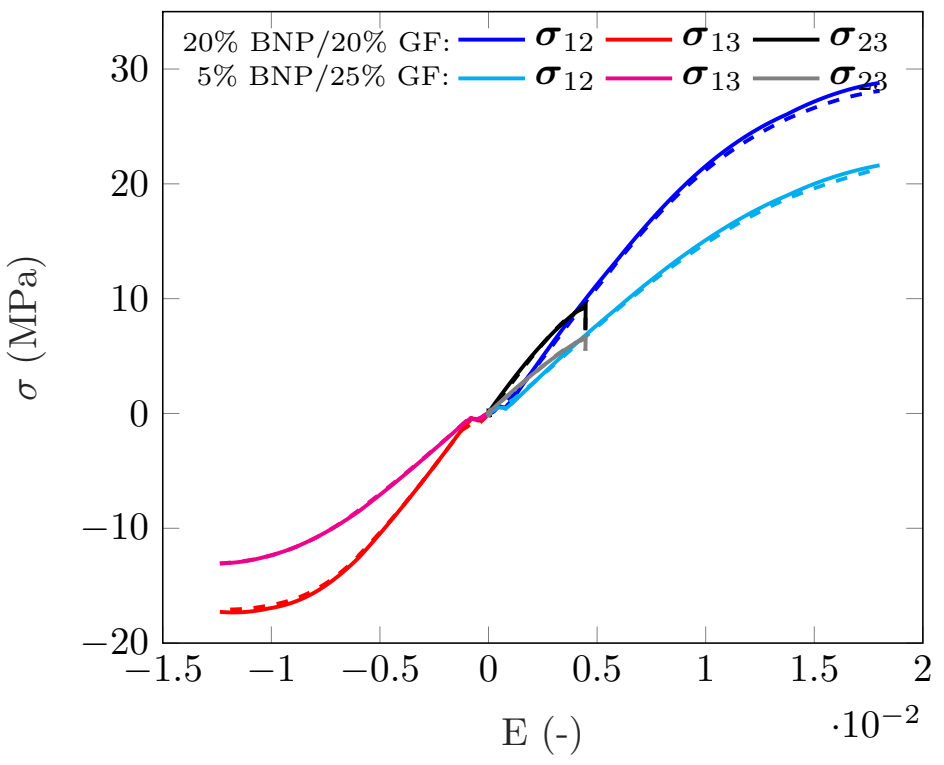

**(b)**

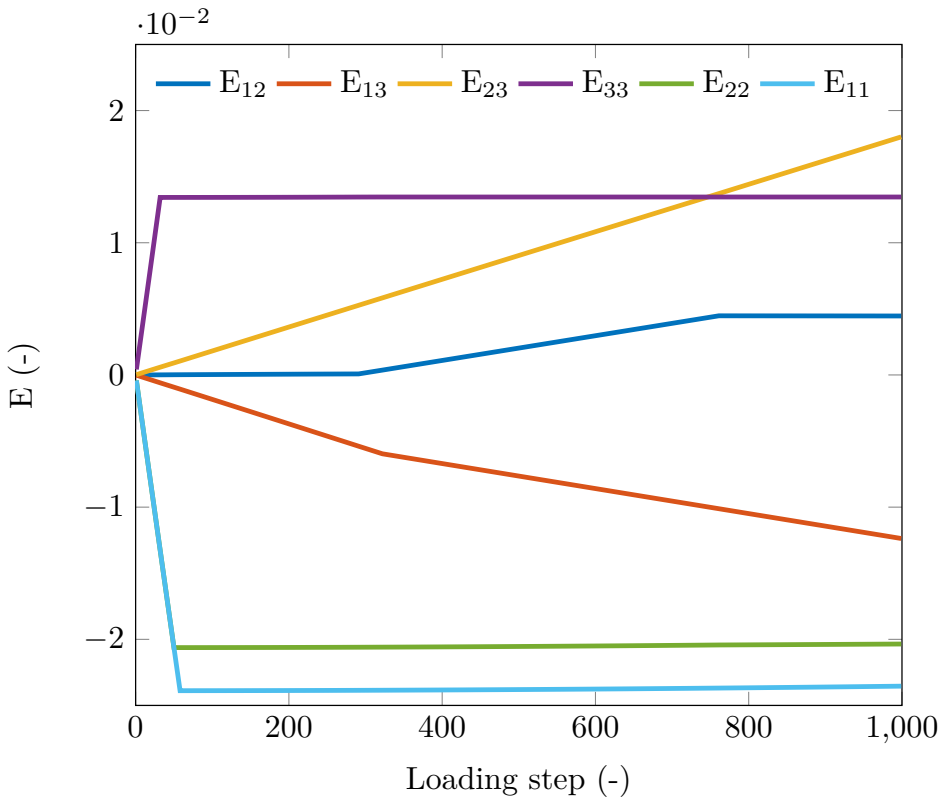

**(c)**

**Figure 7:** The fiber- and nanoparticle reinforced composite under a complex three-dimensional combination of shear and axial loading at dry state. The solid lines represent the classical constitutive model and the dashed lines are predicted by the PIDL model. The diagonal terms of the stress tensor are presented in (a) and the shear terms are shown in (b). The effective strain rate derived from the frobenius norm of the Green strain tensor is $\dot{\varepsilon}_F = 3 \times 10^{-3}\ s^{-1}$. The loading path including all deformation gradient components presented using Eq. (55) as Green strain components.

### *4.3. Modelling of a nanoparticle-filled epoxy nanocomposite with experimental data*

In the following subsection, only experimental data is used for the training and validation of the PIDL model. The accuracy of the deep-learning model in predicting the highly nonlinear behavior of the epoxy system is then evaluated.

*Experiments*

The loading-unloading experiments are performed under various ambient conditions, and the experimental data are then split into training and validation data. For training, data from

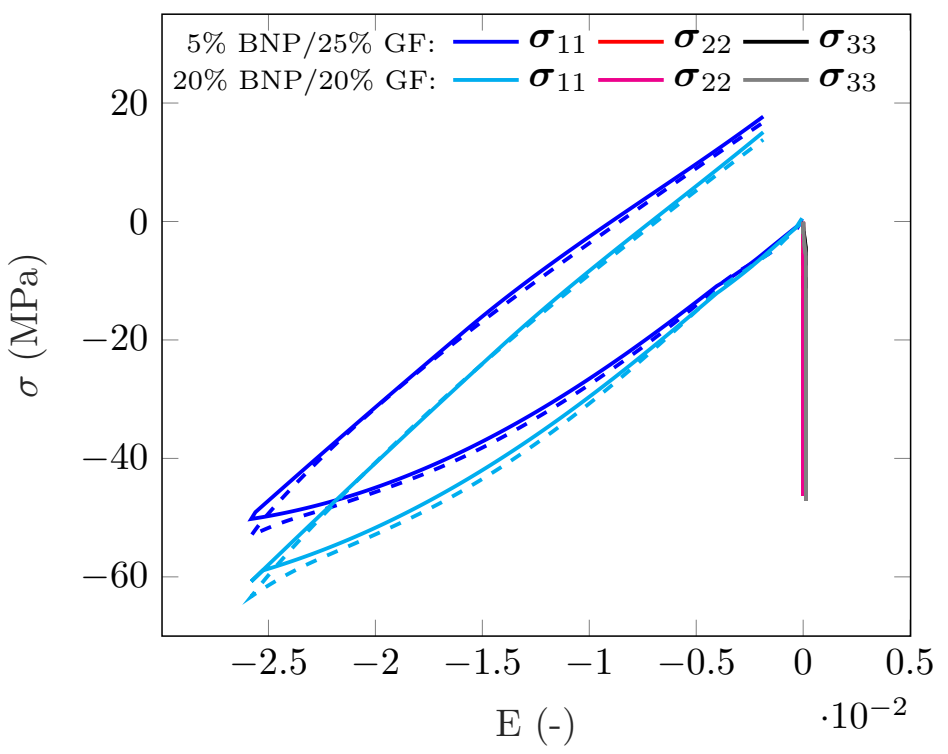

**(a)**

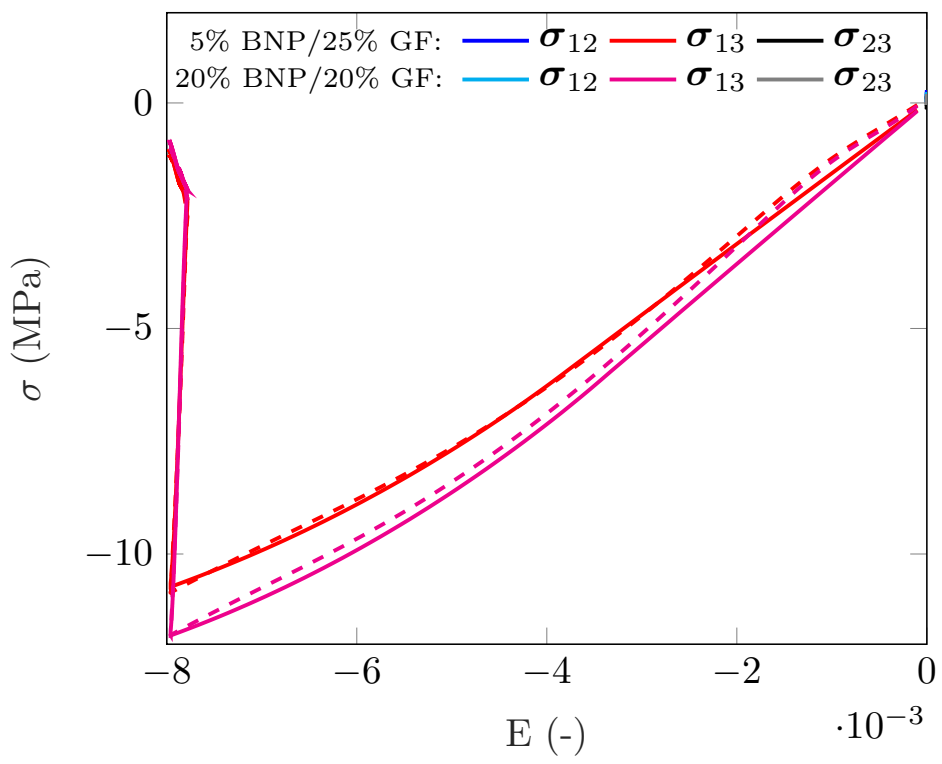

**(b)**

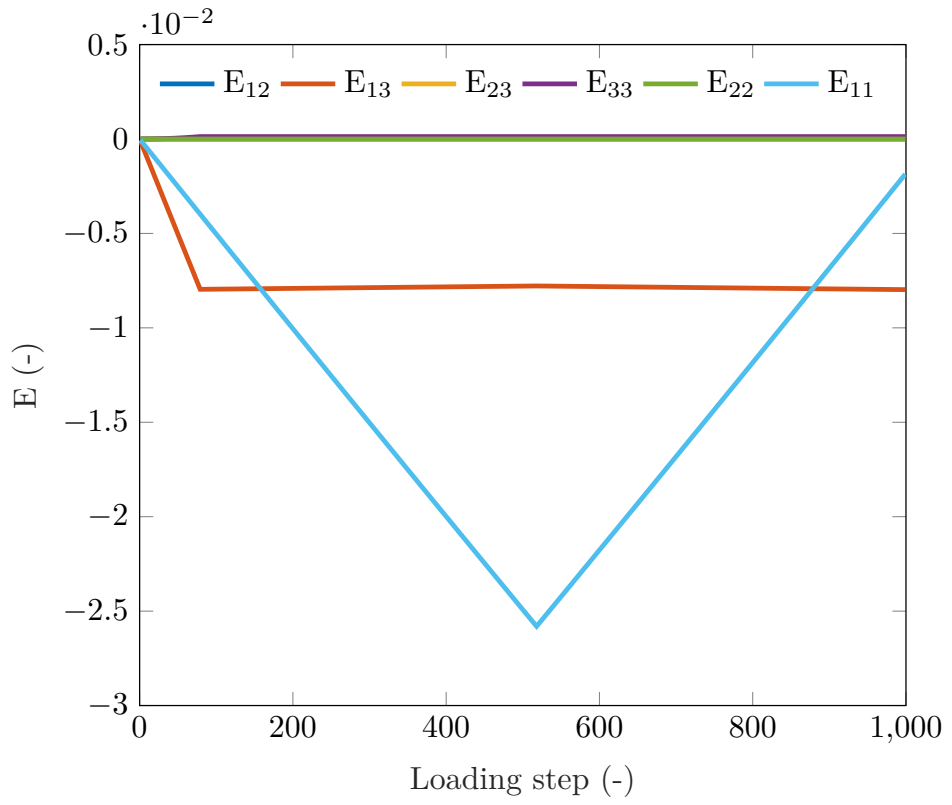

**(c)**

**Figure 8:** The fiber- and nanoparticle reinforced composite under a complex three-dimensional combination of shear and axial loading at saturated state. The solid lines represent the classical constitutive model and the dashed lines are predicted by the PIDL model. The diagonal terms of the stress tensor are presented in (a) and the shear terms are shown in (b). The effective strain rate derived from the frobenius norm of the Green strain tensor is $\dot{\varepsilon}_F = 6 \times 10^{-3}\ s^{-1}$. The loading path including all deformation gradient components presented using Eq. (55) as Green strain components.

loading-unloading experiments with the following conditions are used.

- Nanoparticle volume fraction $v_{np} \in \{0, 10\}\%$.
- Temperature $\Theta \in \{-20, 23, 60\}$°C.
- Moisture content $w_w \in \{0, 1\}$, where 0 stands for the dry state and 1 is the saturated state.

Furthermore, the following data from experiments are used to validate the deep-learning model.

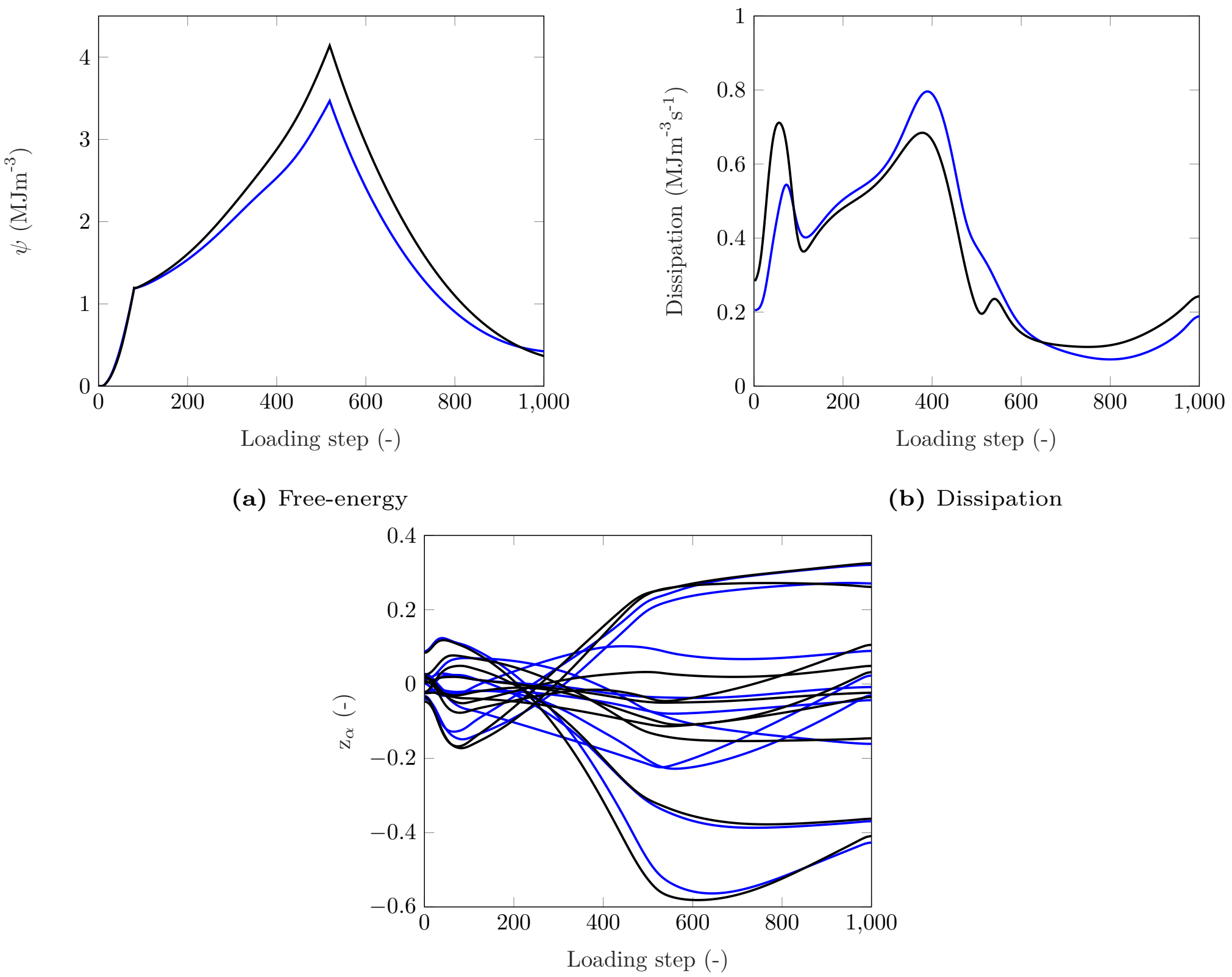

**(a)** Free-energy

**(b)** Dissipation

**(c)** Internal variables

**Figure 9:** Free-energy(a), dissipation(b) and internal variables(c) of the constitutive model as a function of loading steps. The blue lines represent the 5%BNP/25%GF model and black lines represent the 20%BNP/20%GF model.

- Nanoparticle volume fraction $v_{np} \in \{5\}\%$.
- Temperature $\Theta \in \{-20, 23, 50, 60\}$°C.
- Moisture content $w_w \in \{0, 1\}$, where 0 stands for the dry state and 1 is the saturated state.

The specimens for the conditioning and mechanical tests are cut from the panels as presented in Fig. 10. The necking in the tension direction of epoxy systems is a structural instability rather than a material property [9, 12]. Therefore, a notch is inserted to reduce the influence of material imperfections and necking on its yield. The conditioning of the specimens is the same as the previously presented material system. Finally, mechanical loading-unloading tests are produced according to testing standard DIN EN ISO 527-2, using an extensometer to measure the elongation of the specimens and a load rate of 1 mm/min. We apply a cyclic loading to a specific amplitude and unload until the loading force reaches zero. In the next cycle, a higher force than the previous one is applied, and the process is repeated until failure.

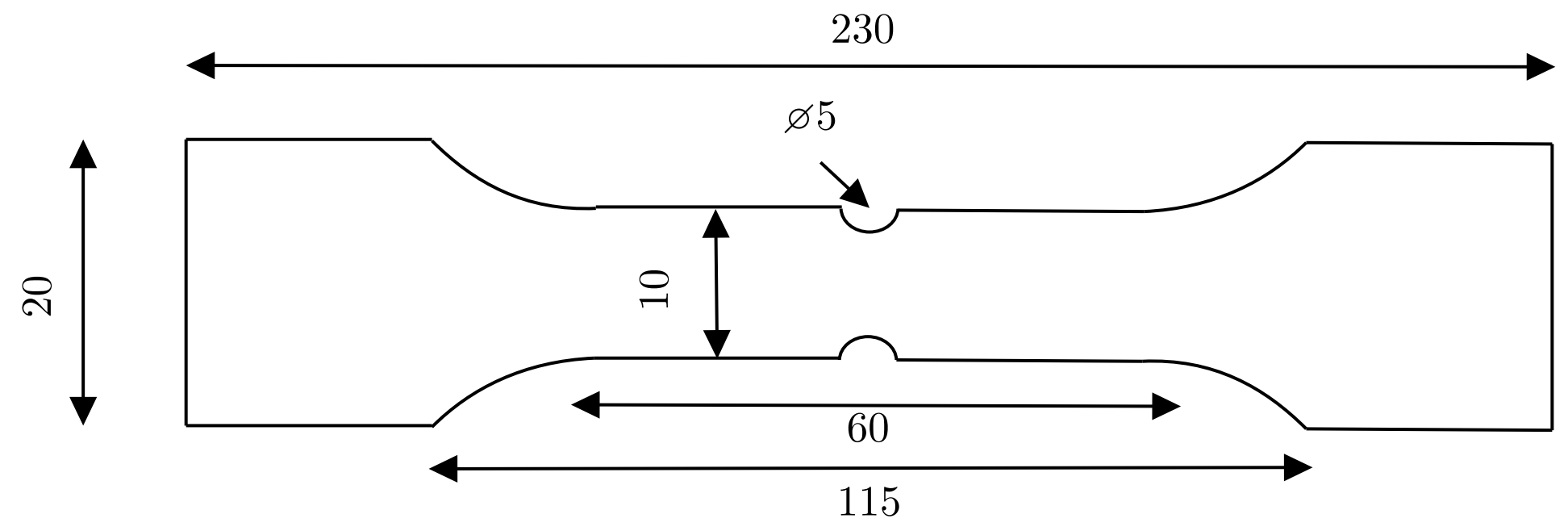

**Figure 10:** Planar dimensions of the specimen for conditioning and mechanical loading-unloading tests with a thickness of 2.3 mm. All dimensions are in millimeters.

*Training data*

Since the nanoparticle-filled epoxy nanocomposite can be regarded as an isotropic material, we include only the first three invariants $I(\mathbf{C}) \in \{\mathrm{I}_1, \mathrm{I}_2, \mathrm{I}_3\}$ to the prediction of the stress tensor in Eq. (9). The hyperparameters for this application case are presented in Table 4.

**Table 4:** Hyperparameters of the PIDL model for the second application case

| Hyperparameter | Value |
|---|---|
| Learning rate $\alpha$ | $5e-4$ |
| Epochs | 5000 |
| Hidden layers ($\mathcal{F}_{lstm}, \mathcal{F}_{znn}, \mathcal{F}_{\psi nn}$) | 2 |
| Number of neurons per hidden layer ($\mathcal{F}_{lstm}, \mathcal{F}_{znn}, \mathcal{F}_{\psi nn}$) | 30 |

Firstly, the number of internal variables is tuned for the best prediction of the data. The training results are shown in Fig. 11. As can be seen, the optimal number of internal variables for predicting stress-strain behavior under the conditions above is eight. Subsequently, we proceed with training our deep-learning model, utilizing eigth internal variables along with the specified hyperparameters.

Fig. 12(a) and (b) illustrate the stress-strain response of the neat epoxy at various temperatures under cyclic loading-unloading conditions. A comparison between the predicted and experimental data demonstrates the deep-learning model's capability to accurately predict

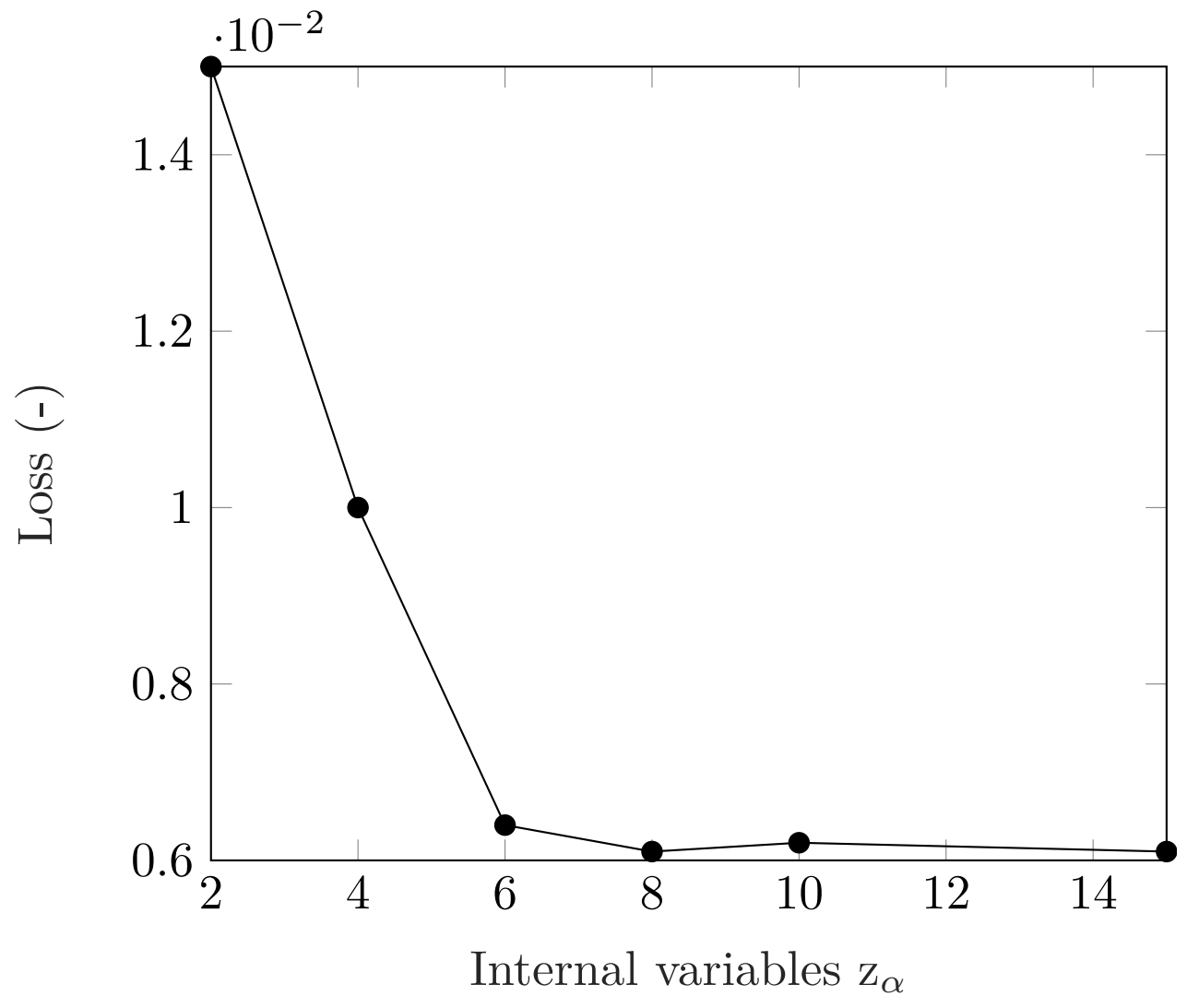

**Figure 11:** Final training loss as a function of internal variables.

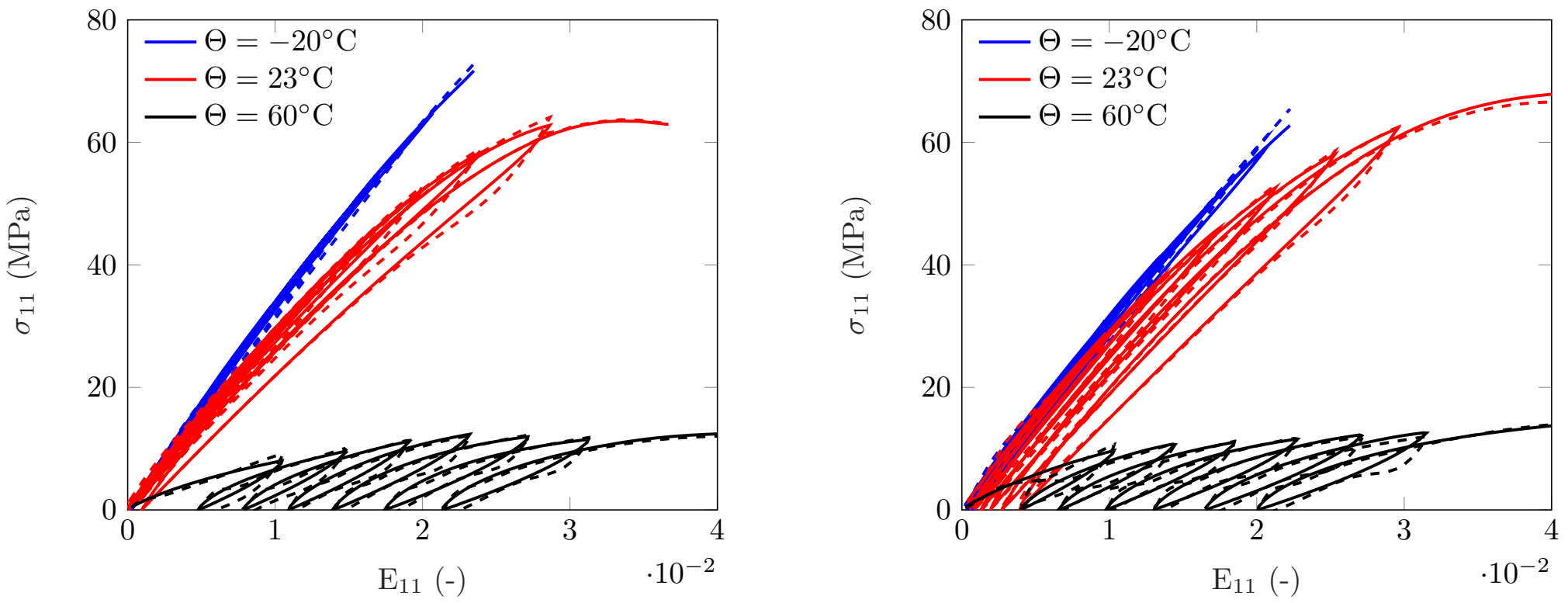

**(a)** Dry state

**(b)** Saturated state

**Figure 12:** Experimental and predicted stress-strain behavior of the PIDL model (dashed lines) for the neat epoxy at different temperatures.

the highly nonlinear viscoelastic behavior under diverse ambient conditions. This accuracy is observed for the 10% BNPs epoxy system too (see Fig. 13).

A notably stiff behavior is observed at lower temperatures of −20 °C, with stiffness decreasing at higher temperatures. Concurrently, increased hysteresis is observed with rising temperature, as anticipated, due to higher molecular chain movement within the polymeric matrix at elevated temperatures [58]. Noteworthy, an increase in strains at zero stress level is observed in the saturated state compared to the dry state at −20 °Cand 23 °C temperatures,

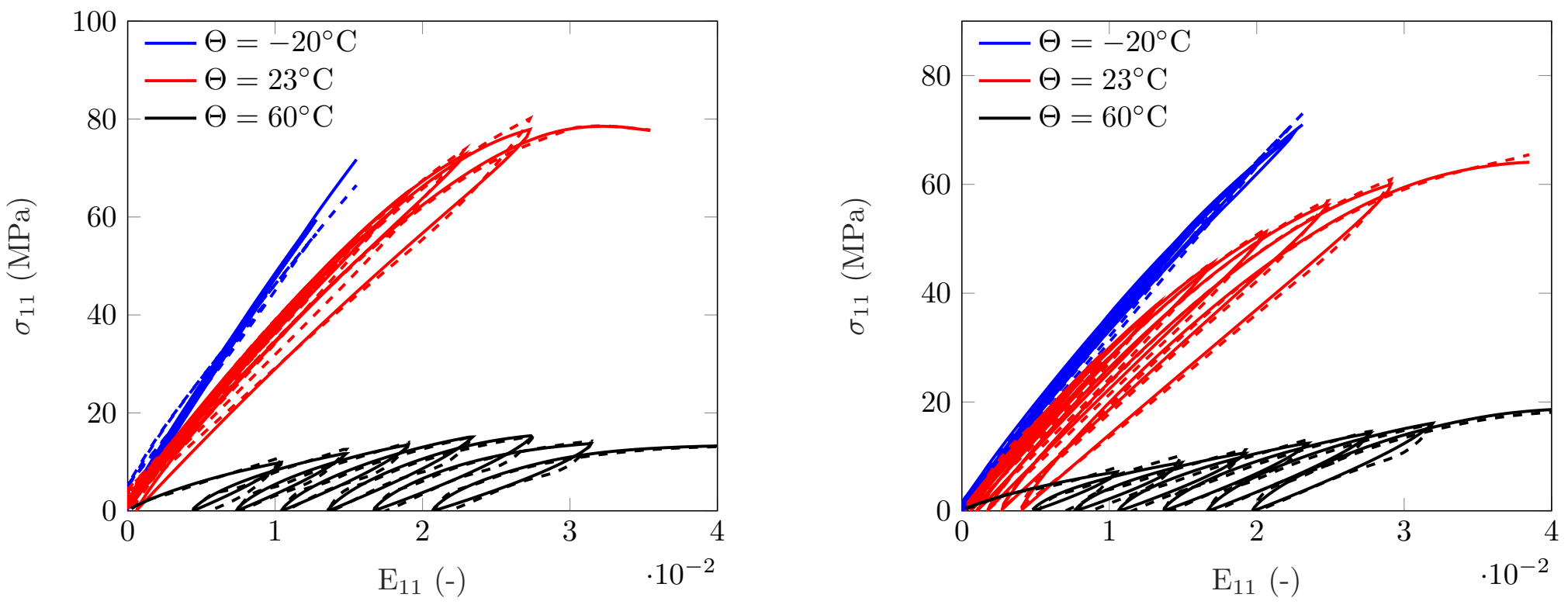

**(a)** Dry state

**(b)** Saturated state

**Figure 13:** Experimental and predicted stress-strain behavior of the PIDL model (dashed lines) for 10% BNPs/nanocomposite at different temperatures.

adding complexity to the prediction of stress-strain behavior. This behavior is not apparent at high temperatures of 60 °C, where the mobility of water molecules is higher [10] and the moisture content decreases rapidly. The PIDL model can learn the stress-strain behavior even for this complex loading-unloading case, yielding accurate results across a wide range of temperatures under dry and saturated conditions. However, some deviations are observed, particularly at high temperatures, where hysteresis and softening increase, and the PIDL model cannot fully depict the experimental observations.

*Validation data*

While the training data is expected to perform well, the PIDL model is validated by using additional data outside the training data for prediction. As stated above, we use data for 5% BNPs volume fraction and add the experimental data at 50 °C to validate the model.

Fig. 14 presents the validation results from experiments and the deep-learning model. An accurate prediction of the PIDL model is observed compared with the experimental data. Noteworthy observations are listed as follows:

- At low temperatures of −20 °C, a stiff behavior is observed. A slightly larger hysteresis can be seen at the saturated state, which underlines the increased mobility of the polymer chains due to the moisture content. The deep-learning model is able to predict the behavior and interpolates successfully between 10% and 0% BNPs volume fraction.
- With temperature increase, the viscoelastic behavior increases as the hysteresis increases.

Also, a larger softening behavior is observed at higher temperatures, which is accurately captured by the PIDL model .

- While the PIDL model accurately interpolates between the BNPs volume fractions, noteworthy interpolation can also be observed at a temperature of 50 °C. The model accurately predicts the peak stresses but tends to overestimate the hysteresis.

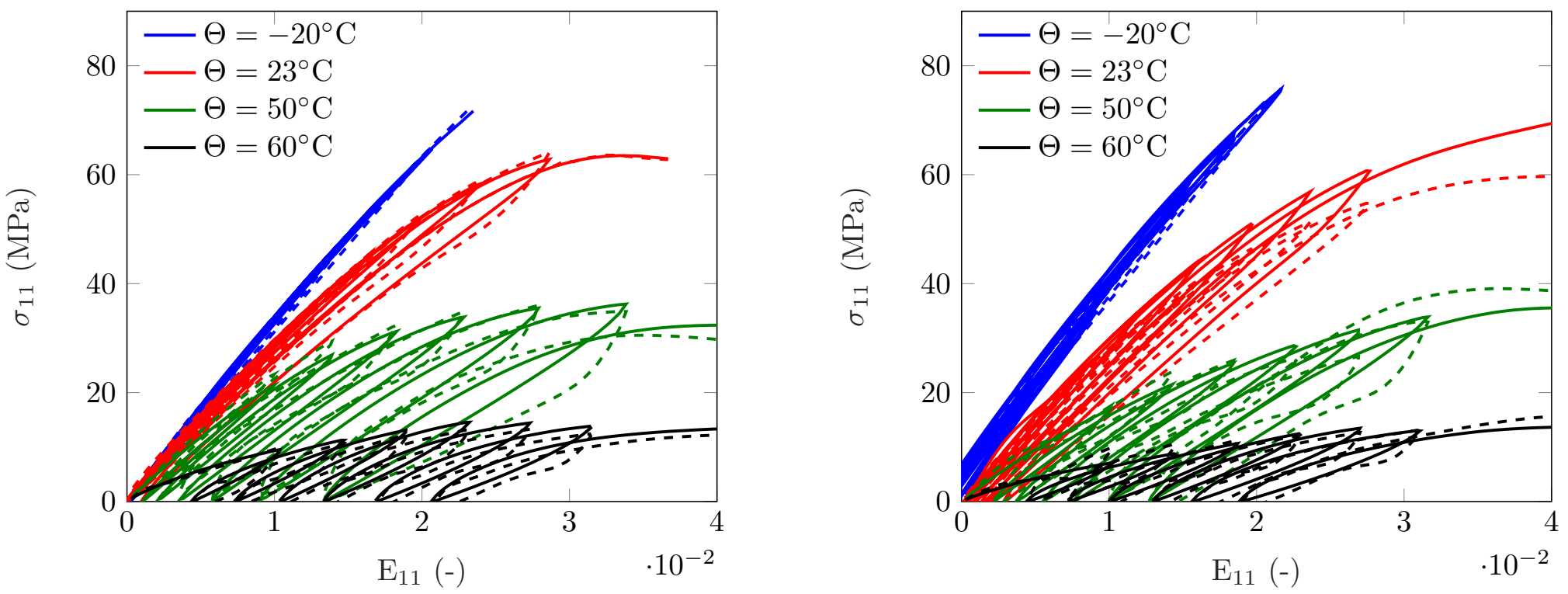

**(a)** Dry state

**(b)** Saturated state

**Figure 14:** Experimental (solid lines) and predicted stress-strain behavior of the PIDL model (dashed lines) for 5% BNPs/nanocomposite at different ambient conditions.

Overall, the PIDL model is able to predict a highly nonlinear stress-strain behavior of the material system at hand under different ambient conditions. To elucidate the behavior in more detailed manner, we present the free-energy, dissipation rate and internal variable behavior during the cyclic loading-unloading paths.

Fig. 15 shows the free-energy evolution as a function of loading steps. Notably, peaks are observed at each loading maxima, which is an expected behavior due to increasing stress and strain. A noticeable complex behavior is seen for the dry and saturated states. At normal room temperature, the amount of free energy is less in a dry state as compared to a saturated state. However, at both low and high temperatures, the free energy levels are similar or higher in a saturated state, which also correlates with the behavior of stress and strain. Maximum energy levels are reached at room temperature, with a subsequent decrease at higher and lower temperatures. The PIDL model can attain the convexity and non-negativity of the free energy, which are crucial to enforcing thermodynamic consistency. Fig. 16(a) to (d) presents the evolution of the internal variables for each simulation case. The first observation, which correlates with the stress-strain behavior, is the scale of the internal variables at different

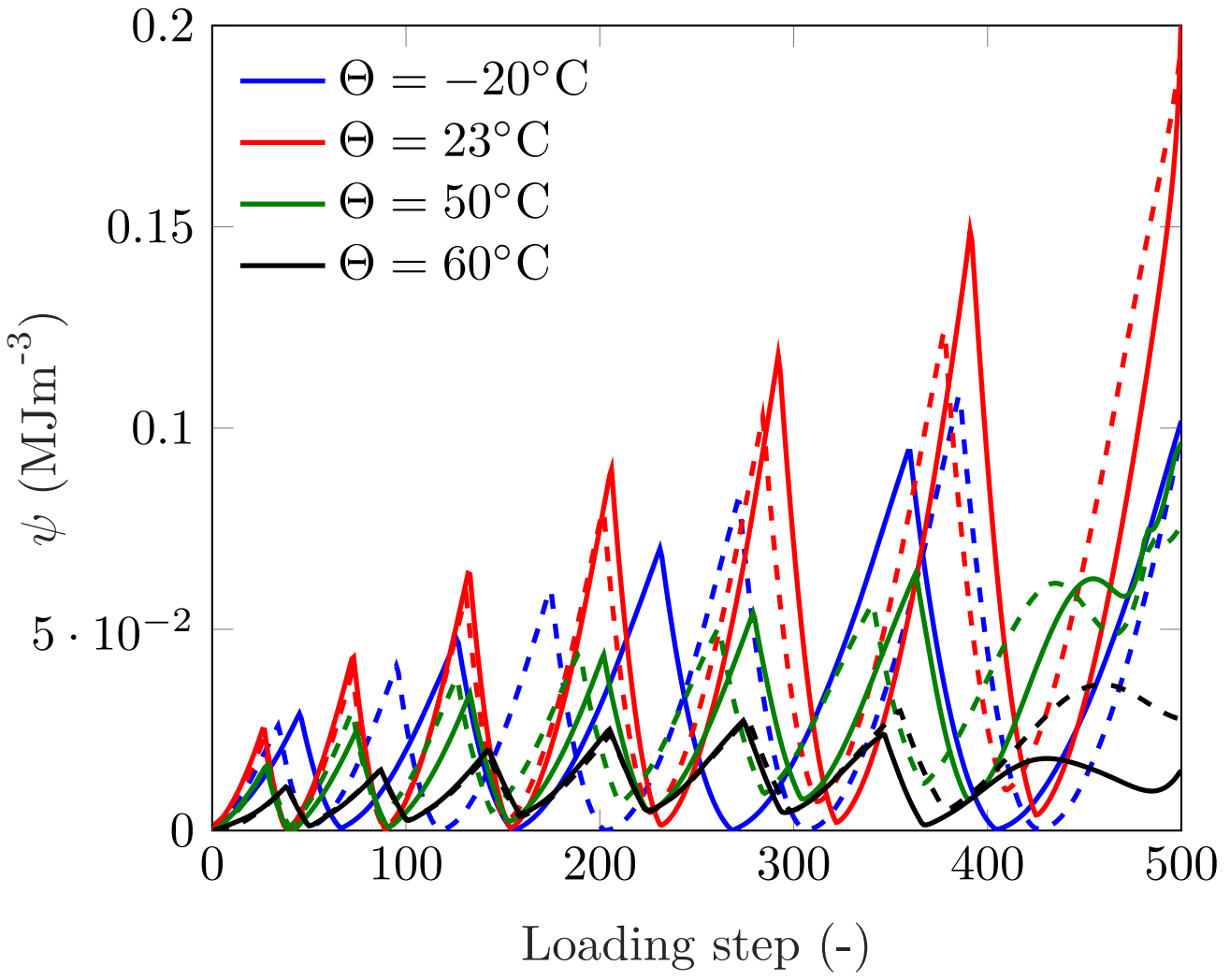

**Figure 15:** Free-energy of the DL model as a function of loading steps. The solid lines represent the dry state and dashed lines correlate to the saturated state

temperatures. The internal variables increase with increasing temperatures indicating an increase in nonlinearities as the temperature increases. The second observation is the increase of the internal variables as the loading continues. Due to higher hysteresis and damage of the material at higher strain, we observe an increase in the internal variables. This correlates with the observations in the experiment because higher strains lead to higher nonlinearities, which are depicted by internal variables. The only exception here is the simulation at $-20\,°C$ (Fig. 16(a)), which also underlines the brittle and almost linear elastic behavior of the stress-strain response. The stress response at these temperature levels does not show an increase of hysteresis with increasing strain, and also, no significant softening behavior can be observed. Since the nonlinear viscoelasticity and softening behavior are implicitly described by internal variables, we do not observe these phenomena at low temperatures, and therefore, the internal variables attain the same levels during the whole deformation path. Thirdly, it should be mentioned that each internal variable shows a highly nonlinear temperature and moisture dependency, underlining the complex nonlinear behavior of the BNP-reinforced nanocomposite. Finally, the evolution of internal variables in the model appears to be mostly symmetric. As one set of internal variables progresses in the positive direction, the complementary subset evolves in a parallel negative direction. This symmetry is indicative of a strong dependency between the internal variables. In Fig. 17, the dissipation of the PIDL model is presented

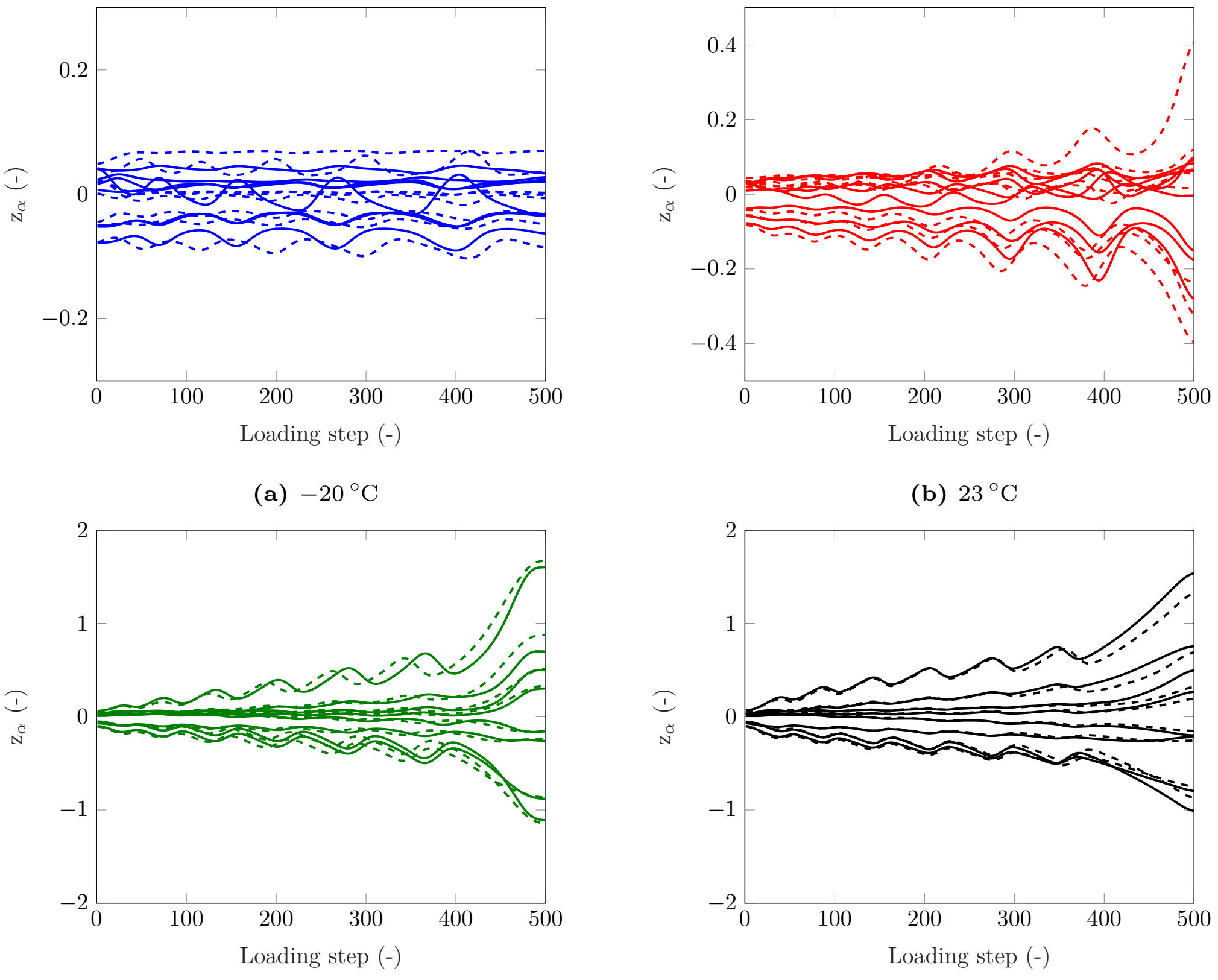

**(a)** −20 °C  **(b)** 23 °C  **(c)** 50 °C  **(d)** 60 °C

**Figure 16:** Evolution of the internal variables $z_\alpha$ as a function of loading steps for different temperatures at dry (solid lines) and saturated (dashed lines) state.

throughout the deformation path. We observe the temperature dependency of the dissipation and a comparable behavior to the evolution of internal variables in Fig. 16. Furthermore, as the deformation increases in Fig. 16, there is a rapid increase in dissipation, which is related to the rise in internal variables as the strains increase. We also observe a low dissipation for the −20 °C, suggesting that the material system exhibits brittle behavior at this temperature. Up to the point of failure, minimal to no nonlinearities are observed, indicating a perfectly elastic material response. The observation suggests that a lower number of internal or no internal variables is probably sufficient to characterize the thermodynamic state of the system at low temperatures. Again, the present PIDL model effectively ensures the positivity of dissipation, thereby enforcing the Clausius-Duhem inequality from Eq. (6).

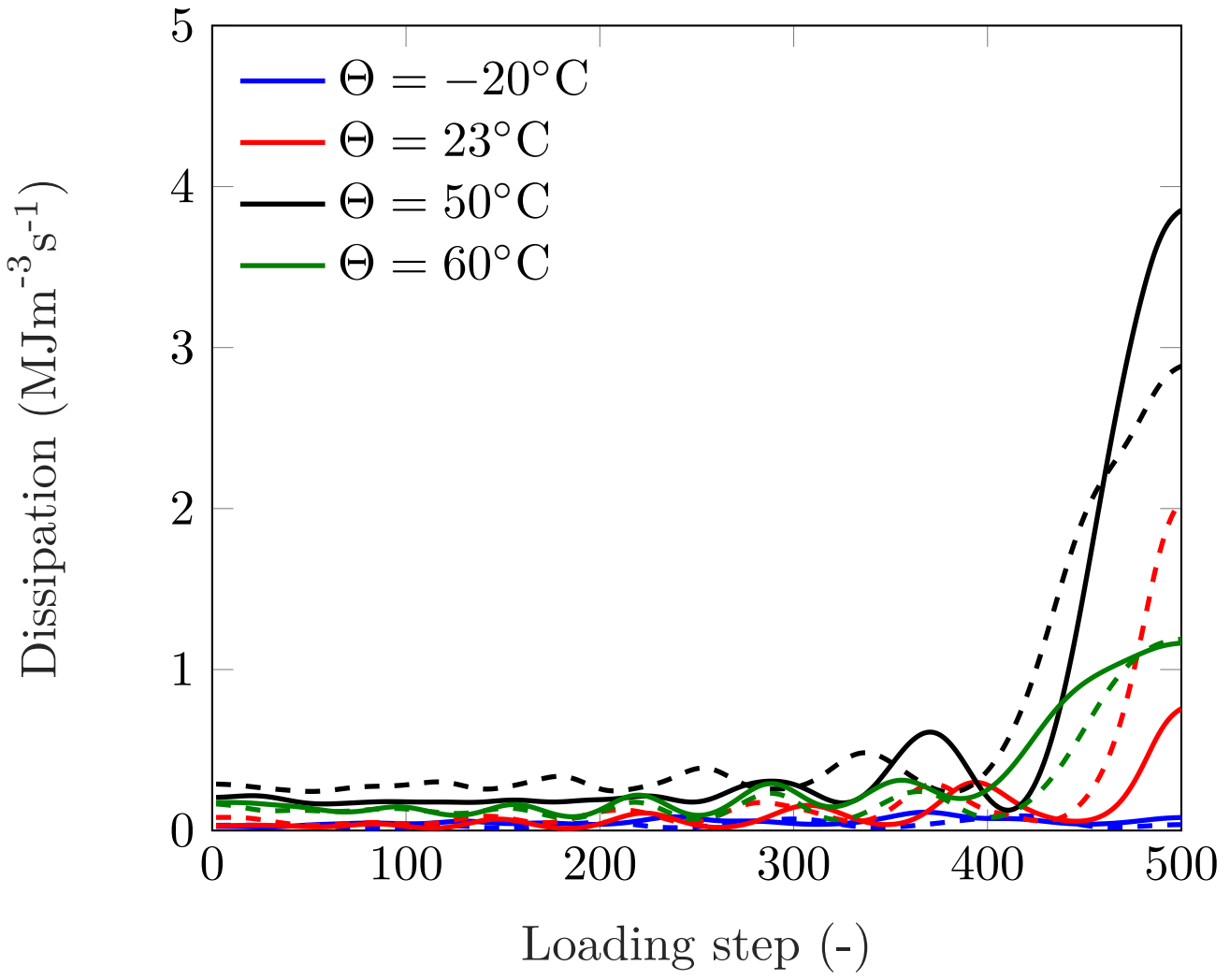

**Figure 17:** Dissipation of the DL model as a function of loading steps. The solid lines represent the dry state and dashed lines correlate to the saturated state

## 5. Summary and conclusions

In this study, a PIDL model has been introduced to investigate the mechanical behavior of fiber reinforced nanoparticle-filled nanocomposites under different ambient conditions at finite deformations. The thermodynamic consistency of the proposed PIDL model has been enforced and several neural networks have been combined to obtain the free energy. The LSTM has been utilized to obtain history informations of the deformation path for this particular material system and a feed-forward neural network has been used to obtain the final internal variables. These serve as an input to another feed-forward neural network together with the invariants of the right Cauchy-Green tensor, leading to the free-energy value. The final stress tensor has been obtained by using automatic differentiation, and the internal dissipation has been calculated using an approximation of the rate of internal variables as presented in Eq. (26).The major results of this study can be summarized as follows:

1. The proposed PIDL model is able to accurately predict the three-dimensional nonlinear behavior of short fiber/epoxy nanocomposites under different ambient conditions. The model is able to sufficiently extrapolate over the training boundaries with a loss of 0.008.
2. Classical constitutive models may not be able to fully capture the complex mechanical behavior of materials under varying temperature and moisture conditions [12, 24]. Since the pass forward of the PIDL model contains almost solely matrix multiplication, with the exception of the automatic differentiation to obtain the stress tensor, the computing times do not vary a lot and are comparable with classical constitutive models. In contrast, the trained PIDL model can accurately capture the highly nonlinear mechanical behavior of nanocomposites subjected to cyclic loading and unloading, taking into account the effects of temperature and moisture content. Moreover, the evolution of the free energy, internal variables, and dissipation are helpful outputs to explain the mechanical behavior. Therefore, the PIDL model can also serve as an explainable neural networks.

While the proposed PIDL model shows effective generalization behavior and leads to accurate results, additional improvements can be made for future work. Firstly, other models have shown that further enforcements of the constitutive conditions can be made [37]. Enforcing these conditions increases the generalization of the model further. Also, additional experiments at different temperatures could lead to better prediction of the hysteresis and softening behavior for the validation data. Especially for the second case, we observe a deviation regarding the hysteresis for the 5%BNPs at 50 °C in Fig. 14, which could be solved if additional temperature ranges are included within the training data. Additionally, an improvement of the PIDL model itself for future works is the implementation of neural integration, which leads to an increase of accuracy for small- and noisy data [59].

In summary, the proposed PIDL model can be further developed. Similar cyclic loading-unloading experiments to the presented ones for simple shear would be helpful to calibrate the model in two- or three-dimensional case, making the PIDL model integration within a finite element framework without the need of synthetic data generation from classical constitutive models a realistic achievement. Due to the enforcement of the constitutive conditions and thermodynamic consistency, the PIDL model is then capable to extrapolate from uniaxial loading to shear or biaxial loading [34]. This will result in a accurate PIDL model completely replacing classical constitutive models.

*Data availability*

The source codes of the PIDL model in this work are available at `https://github.com/BBahtiri/deep_learning_constitutive_model`.

*Acknowledgements*

This work originates from the following research project: "Functionalized, multi-physically optimized Adhesive systems for inherent structural monitoring of rotor blades" ("Func2Ad - Funktionalisierte, multiphysikalisch optimierte Klebstoffsysteme für die inhärente Strukturüberwachung von Rotorblättern"), funded by the Federal Ministry for Economic Affairs and Climate Action, Germany. The authors wish to express their gratitude for the financial support. The authors acknowledge the support of the LUIS scientific computing cluster, Germany, which is funded by Leibniz Universitat Hannover, Germany, the Lower Saxony Ministry of Science and Culture (MWK), Germany and the German Research Council (DFG).

## Appendix A. Derivation of invariants with respect to the right Cauchy-Green deformation tensor

The derivations of the invariants presented in Section 2.2 are calculated as follows:

$$\frac{\partial \mathrm{I}_1}{\partial \mathbf{C}} = \frac{\partial \mathrm{tr}\mathbf{C}}{\partial \mathbf{C}} = \frac{\partial (\mathbf{I} : \mathbf{C})}{\partial \mathbf{C}} = \mathbf{I} \tag{A.1}$$

$$\frac{\partial \mathrm{I}_2}{\partial \mathbf{C}} = \frac{1}{2}\left(2\mathrm{tr}\mathbf{C}\mathbf{I} - \frac{\partial \mathrm{tr}(\mathbf{C}^2)}{\partial \mathbf{C}}\right) = \mathrm{I}_1\mathbf{I} - \mathbf{C} \tag{A.2}$$

$$\frac{\partial \mathrm{I}_3}{\partial \mathbf{C}} = \mathrm{I}_3\mathbf{C}^{-1} \tag{A.3}$$

$$\frac{\partial \mathrm{I}_4}{\partial \mathbf{C}} = \mathbf{a}_0 \otimes \mathbf{a}_0 \tag{A.4}$$

$$\frac{\partial \mathrm{I}_5}{\partial \mathbf{C}} = \mathbf{a}_0 \otimes \mathbf{C}\mathbf{a}_0 + \mathbf{a}_0\mathbf{C} \otimes \mathbf{a}_0 \tag{A.5}$$

$$\frac{\partial \mathrm{I}_6}{\partial \mathbf{C}} = \mathbf{g}_0 \otimes \mathbf{g}_0 \tag{A.6}$$

$$\frac{\partial \mathrm{I}_7}{\partial \mathbf{C}} = \mathbf{g}_0 \otimes \mathbf{C}\mathbf{g}_0 + \mathbf{g}_0\mathbf{C} \otimes \mathbf{g}_0 \tag{A.7}$$

$$\frac{\partial \mathrm{I}_8}{\partial \mathrm{C}} = \frac{1}{2}(\mathbf{a}_0 \cdot \mathbf{g}_0)(\mathbf{a}_0 \otimes \mathbf{g}_0 + \mathbf{g}_0 \otimes \mathbf{a}_0) \tag{A.8}$$